\documentclass[preprint, 5p, times]{elsarticle}

\usepackage{ecrc_RIAI}
\usepackage{subfigure}
\usepackage{url}
\usepackage{float}
\usepackage{subcaption}
\biboptions{numbers,sort&compress}
\setcitestyle{square}
\usepackage{longtable}

\usepackage[english]{babel}  % Language settings
\usepackage[utf8]{inputenc}  % Encoding
\usepackage{amsmath}         % For math environments like \eqref
\usepackage{epstopdf}        % Convert EPS figures to PDF
\usepackage{flushend}        % For equalizing columns on last page

\volume{00}
\firstpage{1}

\runauth{Subrata Kumer Paul et al.}

\jid{RIAI}

\jnltitlelogo{}

\usepackage{amssymb}         % Mathematical symbols

\usepackage[colorlinks=true,linkcolor=blue,citecolor=blue,urlcolor=blue]{hyperref}

\begin{document}

\begin{frontmatter}

\title{Effect of Activation Function and Model Optimizer on the Performance of Human Activity Recognition System Using Various Deep Learning Models}

\author[First]{Subrata Kumer Paul\corref{cor1}\fnref{label2}}
\ead{sksubrata96@gmail.com}

\author[First]{Dewan Nafiul Islam Noor}
\ead{dewannoorcse13@gmail.com}

\author[First]{Rakhi Rani Paul}
\ead{rakhipaul.cse@gmail.com}

\author[Second]{Md. Ekramul Hamid}
\ead{ekram\_hamid@yahoo.com}

\author[Third,Fourth]{Fahmid Al Farid\corref{cor2}}
\ead{fahmid.alfarid@berlinsbi.com}

\author[Fourth]{Hezerul Abdul Karim\corref{cor2}}
\ead{hezerul@mmu.edu.my}
\author[Fifth]{Md. Maruf Al Hossain Prince}
\ead{mah-prince@ieee.org}
\author[Second]{Abu Saleh Musa Miah\corref{cor2}}
\ead{musa.cse@ru.ac.bd}
\address[First]{Department of Computer Science and Engineering, Bangladesh Army University of Engineering \& Technology (BAUET), Qadirabad, Dayarampur, Natore-6431, Bangladesh}
\address[Second]{Department of Computer Science and Engineering, University of Rajshahi, Rajshahi-6205, Bangladesh}
\address[Third]{Faculty of Computer Science and Informatics, Berlin School of Business and Innovation, Karl-Marx-Straße 97-99, Berlin, 12043, Germany}
\address[Fourth]{Centre for Image and Vision Computing (CIVC), COE for Artificial Intelligence, Faculty of Artificial Intelligence and Engineering (FAIE), Multimedia University, Cyberjaya 63100, Selangor, Malaysia}
\address[Fifth]{Department of Computer Science and Engineering, Bangladesh Army University of Science and Technology (BAUST), Saidpur-5220, Bangladesh}

\cortext[cor2]{These authors contributed equally to this work.}
\fntext[label2]{Corresponding author-Email:fahmid.alfarid@berlinsbi.com (Fahmid Al Farid)}
\fntext[label3]{Corresponding author-Email: hezerul@mmu.edu.my (Hezerul Abdul Karim)}
\fntext[label4]{Corresponding author-Email: musa.cse@ru.ac.bd (Abu Saleh Musa Miah)}

\begin{abstract}
Human Activity Recognition (HAR) plays an important role in healthcare, surveillance, and smart environments, where reliable action recognition supports timely decision-making and automation. Although deep learning-based HAR systems are widely adopted, the impact of Activation Functions (AF) and Model Optimizers (MO) on performance has not been sufficiently analyzed, especially in terms of how their combinations influence model behavior in practical scenarios. Most existing studies focus on architecture design, while the interaction between AF and MO choices remains relatively unexplored.
In this work, we investigate the effect of three commonly used activation functions (ReLU, Sigmoid, and Tanh) combined with four optimization algorithms (SGD, Adam, RMSprop, and Adagrad) using two recurrent deep learning architectures, namely BiLSTM and ConvLSTM. Experiments are conducted on six medically relevant activity classes selected from the HMDB51 and UCF101 datasets, considering their suitability for healthcare-oriented HAR applications. Our experimental results show that ConvLSTM consistently outperforms BiLSTM across both datasets. ConvLSTM combined with Adam or RMSprop achieves accuracy levels of up to 99.00\%, demonstrating strong spatio-temporal learning capability and stable performance. While BiLSTM performs reasonably well on UCF101 with accuracy close to 98.00\%, its performance significantly drops to around 60.00\% on HMDB51, indicating limited robustness across datasets and weaker sensitivity to AF and MO variations.These findings highlight that model performance in HAR is strongly influenced by the choice of activation function and optimizer. Based on our analysis, ConvLSTM with ReLU and either Adam or RMSprop emerges as a reliable and effective configuration for accurate human activity recognition. This study provides practical insights for optimizing HAR systems, particularly for real-world healthcare environments where fast and precise activity detection is critical.
\end{abstract}


\end{frontmatter}
%HMDB51 (drink, eat, fall\_floor, sit, stand, walk) and UCF101 (BrushingTeeth, Diving, WritingOnBoard, PushUps, BodyWeightSquats, Typing) 
% \linenumbers

\section{Introduction}
The rapid advancement of Machine Learning (ML) and Deep Learning (DL) has transformed numerous domains, including computer vision, natural language processing, and autonomous systems \cite{Ref1,Ref2,miah_reivew2024methodological,miah2023dynamic_graph_general,miah2024hand_multiculture}. The effectiveness of these models strongly depends on the choice of MO and AF, which govern their learning dynamics and final performance. Optimizers play a critical role in minimizing the loss function and enabling efficient convergence to desirable solutions \cite{Ref3}, while AF introduce non-linearity, allowing neural networks to model complex patterns and relationships \cite{Ref4}. 

However, challenges such as vanishing or exploding gradients, slow convergence, and sensitivity to hyperparameter settings have motivated extensive research into improved optimization algorithms and activation functions to enhance robustness and efficiency \cite{Ref5,Ref6,miah2024sensor,miah2024sign_largescale}. In this context, analysing the interplay between AF and MO is essential for addressing common issues such as low accuracy, unstable training, and computational inefficiency in DL-based systems \cite{Ref6}. By systematically studying the performance of different AF and MO combinations, this work aims to provide insights into their impact on convergence speed, generalization capability, and computational cost. The overall motivation is to develop scalable and robust DL models that can operate reliably in real-world environments, where performance is often constrained by data complexity and limited computational resources \cite{Ref7}. 

Although UCF101 and HMDB51 are among the most widely used datasets in HAR, they exhibit several important limitations \cite{miah2023dynamic_graph_general,hassan2024deep_har_miah}. Both datasets are relatively small and lack the diversity needed to fully capture the complexity of human activities in unconstrained real-world settings \cite{Ref8}. Most videos are collected from curated sources such as YouTube or movies, which introduces a bias towards staged or well-structured environments \cite{Ref9}. As a result, many subtle or less common activities are under-represented. In addition, challenges such as occlusions, variations in viewpoint, background clutter, and changes in illumination make accurate recognition difficult \cite{Ref8}. 

Models also struggle to capture long-term temporal dependencies due to limited temporal depth in many clips \cite{Ref9}. Recent research has shown that even state-of-the-art architectures perform poorly when evaluated on noisy or distorted versions of these datasets \cite{Ref10}. While transformer-based models have achieved encouraging results, particularly in handling temporal sequences and improving robustness to distortions, significant challenges remain in generalizing to unseen actions and adapting to real-time, unconstrained environments \cite{Ref8,Ref9}. The objectives of this research, therefore, include evaluating the effect of AF and MO choices, proposing practical configuration guidelines, and providing empirical evidence to support the selection of appropriate AF and MO combinations for HAR tasks.

The remainder of the paper is organized as follows. Section~2 provides a summary of relevant research. Section~3 describes the datasets used in this work. Section~4 presents the proposed methodology and model configurations. Section~5 reports and analyses the experimental results. Section 6 analyzes the model comparison. Finally, Section~7 concludes the paper and outlines directions for future work.

\subsection{Research Motivation}
HAR has become increasingly significant in domains such as healthcare monitoring, smart surveillance, and context-aware systems \cite{miah2024sensor,miah2024sign_largescale,miah2025methodologica_pd,cmc_har_subrta_miah}. As DL models continue to dominate HAR solutions, their success depends heavily on the appropriate selection of model components, particularly the dataset, AF, and MO. Despite the growing popularity of DL-based approaches, the combined influence of MO and AF on HAR performance remains largely underexplored in the current literature. Most existing studies have concentrated on architectural variations or dataset improvements, often overlooking how fundamental hyperparameters affect learning behaviour and generalization capability. 

Beyond architectural design, HAR performance also depends strongly on hyperparameter choices, in particular the pairing of AFs and MOs. Yet, in most existing studies these components are fixed to conventional combinations (e.g., ReLU with Adam) without systematic comparison. This work is therefore motivated by the need to evaluate commonly used AF and MO pairs on recurrent deep models, using curated subsets of two widely used HAR datasets, and to derive empirically grounded recommendations for healthcare-oriented activity recognition.

HAR plays an essential role in several fields, particularly in medical and healthcare applications \cite{Ref7}. Achieving accurate recognition is challenging due to variations in human movements, environmental conditions, and contextual factors. Human activities can be categorized into several fundamental groups based on purpose, motion characteristics, context, or domain-specific requirements \cite{Ref8}. Furthermore, the percentage distribution illustrated in Figure~\ref{fig:activity_distribution} indicates that research efforts in healthcare-focused HAR remain comparatively limited. This observation further motivates the present study to investigate HAR methods tailored to clinically relevant activities.

\subsection{Major Contributions}
The primary goal of this research is to investigate and quantify the effects of AF–MO combinations on the performance of deep learning-based HAR systems. The main contributions are as follows:
\begin{itemize}
    \item We curate six medically relevant activity classes from each of the HMDB51 and UCF101 datasets, focusing on movements and self-care tasks commonly used in mobility assessment, fall detection, and daily activity monitoring.
    \item We systematically evaluate three activation functions (ReLU, Sigmoid, Tanh) and four optimizers (SGD, Adam, RMSprop, Adagrad) across two recurrent architectures, ConvLSTM and BiLSTM, for video-based HAR.
    \item We show that ConvLSTM, when paired with ReLU or Sigmoid and adaptive optimizers such as Adam or RMSprop, achieves up to 99\% accuracy on both subsets, while BiLSTM performs strongly on UCF101 but struggles on HMDB51, highlighting dataset-dependent behaviour.
    \item Unlike prior work that mainly introduces new architectures, this study explicitly analyses how AF and MO choices influence convergence and accuracy, and provides practical guidelines for selecting suitable configurations in healthcare-oriented HAR scenarios based on public video datasets.
\end{itemize}

These findings provide actionable recommendations for configuring recurrent deep-learning models and support researchers and practitioners in building effective and reliable human activity recognition systems.

\section{Related Works}
Numerous studies have explored the application of  DL techniques for HAR. Earlier HAR methods relied heavily on handcrafted spatial and temporal features, which often produced suboptimal performance because of variations in human movements, camera angles, illumination, and background clutter \cite{Ref1,akash2024_twohar_miah,Naj_ICECC2024_miah_har_conf,shin2025_HAR_surv_miah}. With the advancement of computer vision, CNN-based models have demonstrated strong capability in extracting discriminative spatial features, leading to significant improvements in HAR tasks \cite{Ref2}. Recurrent neural networks, including LSTM, BiLSTM, and ConvLSTM, further enhanced HAR performance by learning long-term temporal dependencies and motion dynamics in video sequences \cite{Ref3}. More recently, transformer-based architectures such as ViT have been incorporated into HAR due to their ability to capture long-range spatio-temporal relationships.
\begin{figure}[ht]
    \centering
    \includegraphics[width=1.0\linewidth]{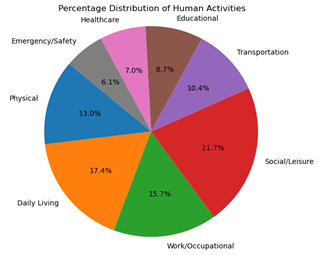} % <-- Replace image file name
    \caption{Percentage distribution of different types of human activities}
    \label{fig:activity_distribution}
\end{figure}
Figure~\ref{fig:activity_distribution} presents the percentage distribution of different types of human activities, highlighting the growing relevance of HAR in healthcare applications.

A number of recent studies related to HAR are summarized in Table~\ref{tab:relatedworks}. These works investigate diverse architectures ranging from CNN–RNN hybrids and lightweight mobile models to transformer-based encoders. Although these models achieve competitive performance, several consistent limitations can be observed:

% \begin{itemize}
%     \item \textbf{Limited exploration of activation functions.} Most studies predominantly rely on ReLU, with occasional use of Sigmoid or Tanh, without systematically comparing alternative activation functions and their effect on sequence models.

%     \item \textbf{Optimizer choices rarely analyzed.} Adam is widely used, while optimizers such as RMSprop, SGD, and Adagrad are often employed without evaluating their comparative performance or convergence behaviour in HAR contexts.

%     \item \textbf{Architectural focus rather than hyperparameter analysis.} Existing works mainly emphasize introducing new architectures or fusion strategies, while comparatively little attention is given to understanding how fundamental hyperparameters (AF and MO) affect HAR model performance.

%     \item \textbf{Real-time and robustness gaps.} Many works achieve high accuracy in controlled settings but do not evaluate robustness under noisy or unconstrained real-world conditions, which is critical for healthcare, surveillance, and assistive environments.
% \end{itemize}

The consistent reliance on specific activation functions and optimizers indicates a gap in the literature: the interaction between AF and MO is rarely studied, even though these hyperparameters significantly influence model convergence, robustness, and generalization. This gap motivates the present study, which systematically evaluates the performance of multiple AF and MO combinations across recurrent architectures (ConvLSTM and BiLSTM) using medically relevant activity subsets from HMDB51 and UCF101. Our goal is to offer practical guidance for selecting appropriate activation–optimizer pairs when designing efficient, stable, and application-ready HAR systems.

\begin{table*}[ht]
\centering
\tiny
\caption{Summary of recent HAR studies and their research gaps}
\label{tab:relatedworks}
\begin{tabular}{p{0.7cm} p{1.2cm} p{1.7cm} p{3.3cm} p{1.8cm} p{1.6cm} p{1.5cm} p{3.5cm}}
\hline
\textbf{Ref} & \textbf{DOI Year} & \textbf{Dataset} & \textbf{Methods} & \textbf{Activation Functions} & \textbf{Optimizer} & \textbf{Accuracy} & \textbf{Research Gap} \\
\hline
{\cite{Ref11}} & 2025 & KTH, UCF Sports, HMDB51 & 2D Conv-RBM + LSTM + optimized frame selection & ReLU (Conv-RBM), Sigmoid (LSTM) & Adam & 97.3\%, 94.8\%, 81.5\% & Limited to static backgrounds; lacks real-time processing. \\

{\cite{Ref12}} & 2024 & KTH, UCF Sports, HMDB51 & CLIP + temporal prompt reparam + dual semantic supervision & ReLU, GELU & Adam & 97.3\%, 94.8\%, 81.5\% & High computation cost; limited robustness in low-resource languages. \\

{\cite{Ref13}} & 2024 & KTH, HMDB51 & Review of CNNs, RNNs, Transformers & ReLU, Softmax & SGD, Adam & 96.3\% & Needs real-time efficiency, cross-domain adaptation. \\

{\cite{Ref14}} & 2024 & SisFall, PAMAP2 & Wavelet + CNN, MobileNetV3, ResNet, GoogleNet & ReLU & Adam & --, 98.1\% & Limited wavelet exploration; not focused on real-time. \\

{\cite{Ref15}} & 2024 & MeltdownCrisis & ViT + ResNet hybrid & ReLU & Adam & 92\% & No comparison with autism-specific methods. \\

{\cite{Ref16}} & 2023 & HMDB51, UCF101, NTU & Mobile-CNNs + ViT & ReLU, Hard-Swish, SiLU & SGD, Adam & 91.61–95.34\%  & Needs real-world robustness testing. \\

{\cite{Ref17}} & 2022 & V1, Jester, UCF101, HMDB51 & MEST: Spatio-Temporal Encoder model & ReLU & Adam & Competitive & Limited real-time scalability. \\

{\cite{Ref18}} & 2021 & HWU-USP & CNN + LSTM fusion & ReLU, Sigmoid, Tanh & Adam & 81.5\% & Weak temporal reasoning; lacks scalability. \\
\hline
\end{tabular}
\end{table*}

\section{Datasets Description}
\begin{figure*}[htbp]
    \centering
    \setlength{\tabcolsep}{4pt} % column gap ( চাইলে কম/বেশি করতে পারো )

    \begin{tabular}{cc}
        % ---------- Row 1 ----------
        \includegraphics[width=0.45\textwidth]{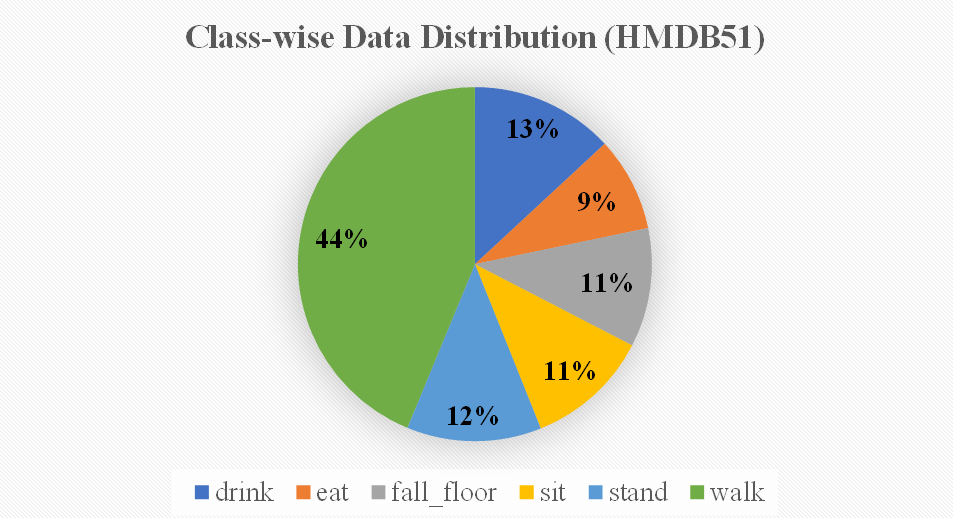} &
        \includegraphics[width=0.45\textwidth]{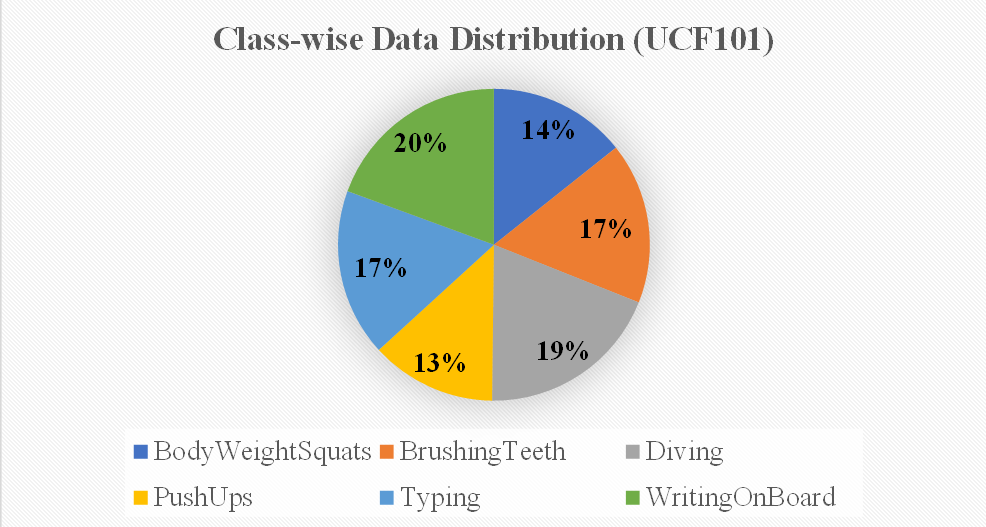} \\
        (a) Class-wise data distribution (HMDB51) &
        (b) Class-wise data distribution (UCF101) \\[0.4cm]

        % ---------- Row 2 ----------
        \includegraphics[width=0.45\textwidth]{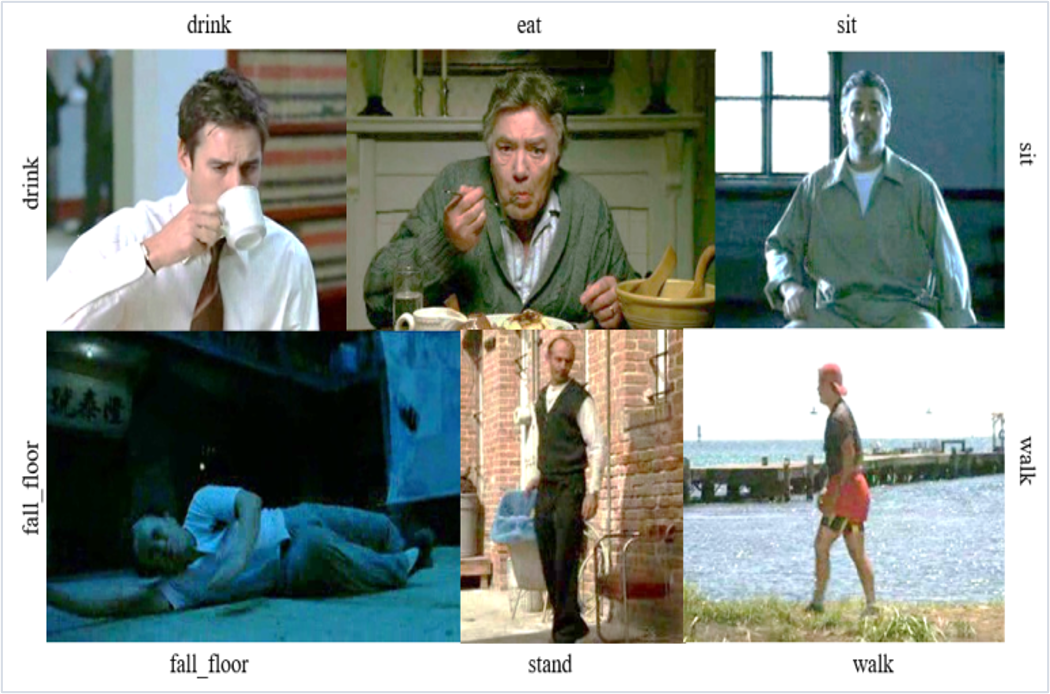} &
        \includegraphics[width=0.45\textwidth]{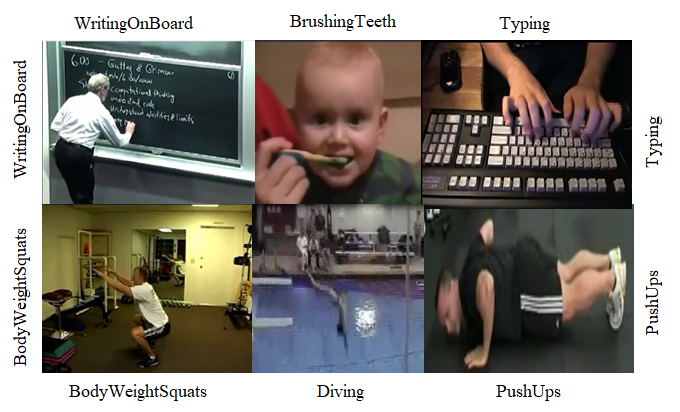} \\
        (c) Sample frames from selected HMDB51 classes &
        (d) Sample frames from selected UCF101 classes \\
    \end{tabular}

    \caption{Overview of dataset characteristics}
    \label{fig:data_distribution}
\end{figure*}

The quality and relevance of the dataset play a crucial role in the performance of any machine learning or deep learning model, including those used in HAR \cite{Ref19}. Therefore, selecting appropriate datasets based on the goals of the experiment is essential. In this study, we consider two widely used public datasets: HMDB51 and UCF101.
HMDB51 is a well-known dataset consisting of 51 human action categories. Each category contains at least 101 video clips collected from diverse sources such as movies, YouTube, and other online platforms. The videos are recorded at 24 FPS, and the dataset contains a total of 6,766 clips in \texttt{.mp4} format, with a total size of approximately 8 GB. For this research, we focus on six medically relevant activities: \textit{fall\_floor}, \textit{walk}, \textit{stand}, \textit{eat}, \textit{sit}, and \textit{drink}. These actions are commonly used in healthcare applications involving mobility assessment and activity monitoring. Video durations range from approximately 3 to 15 seconds \cite{Ref20}.
UCF101 is another widely used HAR dataset containing 13,320 video clips across 101 categories. The videos are collected from YouTube and represent a wide range of human activities recorded in realistic, unconstrained environments. Each video is encoded in \texttt{.avi} format at an average frame rate of 24 FPS. For this study, six medically relevant classes were selected: \textit{BrushingTeeth}, \textit{Diving}, \textit{WritingOnBoard}, \textit{PushUps}, \textit{BodyWeightSquats}, and \textit{Typing}. These activities represent ADL-related movements and self-care tasks, making them suitable for healthcare monitoring scenarios \cite{Ref21}. Figure~\ref{fig:data_distribution} shows the class-wise distribution of both datasets, and Table~\ref{tab:distribution} presents the distribution of videos across each selected class.It is worth noting that the selected HMDB51 subset exhibits class imbalance (for example, the \textit{walk} class contains substantially more clips than the others). In this work, we rely on shuffled mini-batch training without explicit rebalancing and primarily report overall accuracy. A more detailed treatment of class imbalance and its effect on per-class performance is left for future work.

\begin{table}[ht]
\centering
\caption{Distribution of videos in each category}
\label{tab:distribution}
\begin{tabular}{l r l r}
\hline
\textbf{HMDB51 Dataset} & & \textbf{UCF101 Dataset} & \\
\hline
drink & 164 & BodyWeightSquats & 112 \\
eat & 108 & BrushingTeeth & 131 \\
sit & 142 & Diving & 150 \\
stand & 154 & PushUps & 102 \\
walk & 548 & Typing & 136 \\
fall\_floor & 136 & WritingOnBoard & 152 \\
\hline
\end{tabular}
\end{table}

In this study, we specifically chose these six classes from each dataset because they are directly relevant to healthcare-oriented HAR applications. Activities such as walking, sitting, standing, and falling are fundamental for fall detection, rehabilitation tracking, and elderly mobility assessment. Self-care tasks like eating, brushing teeth, and writing are commonly used in ADL evaluations and cognitive or motor-function monitoring \cite{Ref22,Ref23}. In contrast, sports-oriented classes from the original datasets were excluded due to high variability, limited relevance in medical contexts, and reduced interpretability in clinical environments \cite{Ref24}. Restricting the analysis to medically relevant classes also alleviates issues related to class imbalance and label noise. Prior work has shown that domain-specific class selection improves model accuracy and makes HAR systems more suitable for real-world healthcare scenarios, particularly in edge devices or low-resource environments \cite{Ref23,Ref25}.
The dataset preparation pipeline begins with extracting frames at 24 FPS from each video. All frames are resized to 64$\times$64 pixels with three RGB channels, normalized, and encoded before splitting into training, validation, and testing sets. This ensures consistent feature extraction and promotes stable model training \cite{Ref26}. The video clips do not require low-level preprocessing or enhancement, making them suitable for direct training of LSTM-based architectures \cite{Ref27}. Video durations range from 0.73 to 17.66 seconds across both datasets, and all frames are stored as \texttt{.jpg} images to facilitate efficient processing.
These six classes were chosen because they are clinically meaningful for healthcare HAR tasks such as fall detection, mobility assessment, and daily activity monitoring.

\section{Proposed Methodology}
\begin{figure}[ht]
    \centering
    \includegraphics[width=1.0\linewidth]{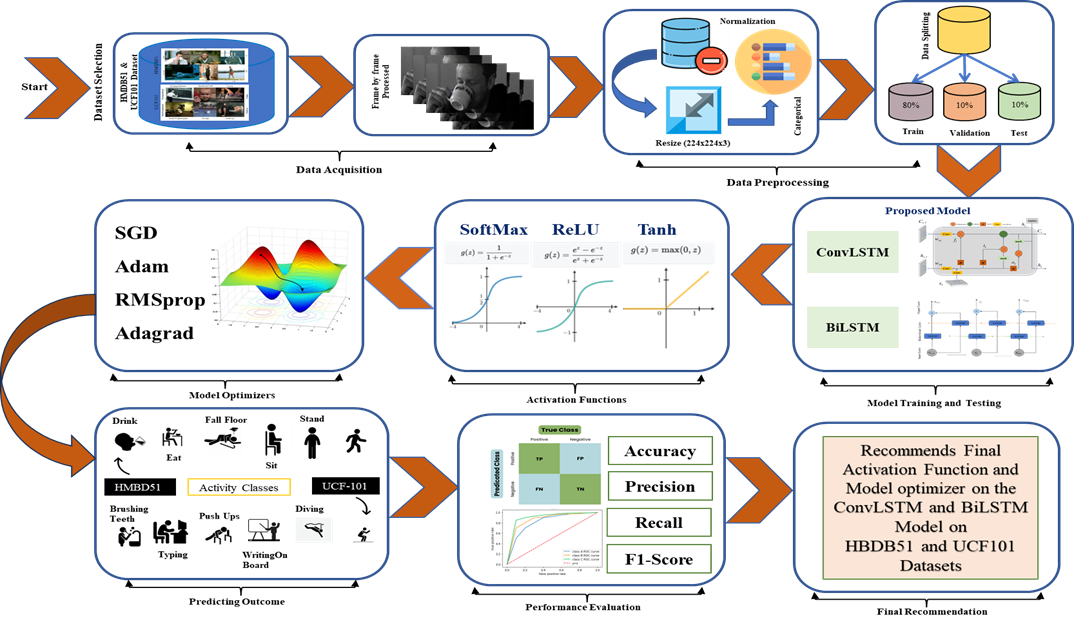} % <-- Replace image file name
    \caption{The overview of Proposed Methodology}
    \label{fig:methodology}
\end{figure}

Figure~\ref{fig:methodology} illustrates the overall workflow of the proposed HAR framework. From Figure \ref{fig:data_distribution}, the process begins with the collection of video data showing six activities from both datasets. Videos are uniformly sampled into fixed-length frame sequences, preprocessed, and split into training, validation, and testing sets. In our study, two different models are used to classify the activities. These models are evaluated by pairing them with four different MOs and three AFs. This work reports overall accuracy, per-class accuracy, and confusion matrices to evaluate classification performance. To ensure fair evaluation, a video-wise split was used so that frames from the same video never appear in both training and testing sets, eliminating any possibility of data leakage. To ensure a fair evaluation, a video-wise split was used so that frames from the same video never appear in both training and testing sets, eliminating any possibility of data leakage. All experiments were run multiple times with different random seeds to verify that the observed trends were stable. For each configuration, we monitored the training and validation accuracy and loss curves to assess convergence behaviour. 

Then the model's performance is evaluated using metrics such as Accuracy (A), Precision (P), Recall (R), and possibly the ROC curve to assess its effectiveness. Along with accuracy, precision, recall, and F1-score were computed to provide a balanced evaluation, especially in the presence of class imbalance. The model's performance is primarily evaluated using classification accuracy on the held-out test sets. During training we also monitored the evolution of training and validation loss and accuracy to check for overfitting and convergence behaviour.

\subsection{Deep Learning Models}
Traditional machine learning methods in HAR often rely on hand-crafted features, which can be labor-intensive, dataset-specific, and limited in capturing the complex temporal and spatial dynamics of human activities. In contrast, deep learning models have emerged as powerful tools that significantly enhance the performance and robustness of HAR systems. The study investigates the following two deep learning architectures to analyze the impact of activation functions and optimizers on HAR performance. In this work, we employed BiLSTM and ConvLSTM architectures instead of heavier backbones of other DL models due to their superior ability to capture both spatial and long-term temporal dependencies inherent in video-based HAR tasks. The ConvLSTM model, in particular, was shown to outperform traditional LSTM by explicitly modeling spatiotemporal correlations leading to enhanced performance in activity recognition scenarios \cite{Ref28},\cite{Ref29}. LSTM based models effectively extract long-term motion patterns across video frames, whereas pure CNN architectures often struggle to capture this temporal context \cite{Ref30},\cite{Ref31}. While some DL models excel in global spatial dependencies, they introduce significantly higher computational complexity and require larger training data, making them less suitable for medical HAR tasks where real-time inference and resource constraints are critical \cite{Ref31},\cite{Ref32}. Our selection of both models strikes an ideal balance offering robust spatiotemporal modeling capabilities while maintaining low latency, simpler training, and effective performance on limited medical datasets. This design choice aligns with prior research emphasizing their applicability in healthcare-oriented HAR, particularly for fall detection, rehabilitation monitoring, and daily activity analysis.

\subsubsection{ConvLSTM (Convolutional Long Short-Term Memory)}
\begin{figure}[ht]
    \centering
    \includegraphics[width=1.0\linewidth]{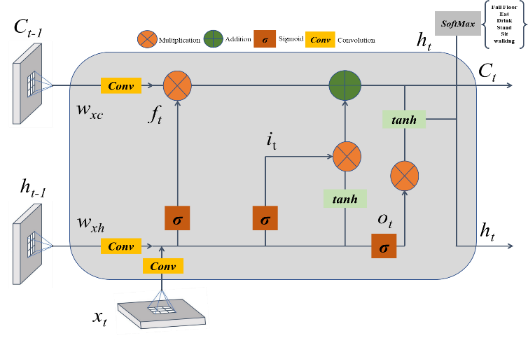} % <-- Replace image file name
    \caption{The internal architecture of ConvLSTM}
    \label{fig:convlstm_arch}
\end{figure}

ConvLSTM is particularly useful for tasks involving video analysis, action recognition, and other applications that require modeling the temporal evolution of spatial data. The ConvLSTM (Convolutional Long Short-Term Memory) architecture extends the traditional LSTM by incorporating convolutional operations to preserve spatial structure while learning temporal dependencies \cite{Ref33}. Figure \ref{fig:convlstm_arch} illustrates the control-flow representation of the ConvLSTM, showing the input gate ($i_t$), forget gate ($f_t$), output gate ($o_t$), and the cell state ($C_t$). The cell state $C_t$ is updated using the input and forget gates along with the candidate cell state, as described in the standard ConvLSTM formulation \cite{Ref33,Ref34}. The equations for each gating mechanism are provided below.

\begin{equation}
i_t = \sigma \left( W_{ii} * X_t + W_{hi} * h_{t-1} + b_{ii} + b_{hi} \right)
\label{eq:conv_i}
\end{equation}

\begin{equation}
f_t = \sigma \left( W_{if} * X_t + W_{hf} * h_{t-1} + b_{if} + b_{hf} \right)
\label{eq:conv_f}
\end{equation}

\begin{equation}
o_t = \sigma \left( W_{io} * X_t + W_{ho} * h_{t-1} + b_{io} + b_{ho} \right)
\label{eq:conv_o}
\end{equation}

\begin{equation}
C_t = f_t \odot C_{t-1} + i_t \odot \tanh\left( W_{ic} * X_t + W_{hc} * h_{t-1} + b_{ic} + b_{hc} \right)
\label{eq:conv_c}
\end{equation}

In this work, the ConvLSTM model is designed to understand both what is happening in each video frame and how those events evolve over time. Each clip is uniformly sampled into 30 frames (64×64×3), normalized, and fed into a two-layer ConvLSTM network. Unlike traditional LSTM, ConvLSTM applies convolutional filters at every time step, enabling the detection of spatial patterns such as shapes, edges, and motion regions, while the LSTM gating mechanism preserves temporal continuity across frames.The first ConvLSTM2D layer extracts local spatial features and short-term motion patterns, whereas the second ConvLSTM2D layer builds richer spatiotemporal representations by linking these features across a longer temporal span. The resulting feature maps are then flattened and passed through fully connected layers, culminating in a Softmax classifier that outputs probabilities for the six healthcare-related activity classes.

\begin{equation}
h_t = o_t \odot \tanh(C_t)
\label{eq:hidden_state}
\end{equation}

Here,  
$W_{ii}, W_{if}, W_{io}, W_{ic}$ are weight matrices for input-to-gate connections,  
$W_{hi}, W_{hf}, W_{ho}, W_{hc}$ are weight matrices for hidden-to-gate connections,  
and $b_{ii}, b_{if}, b_{io}, b_{ic}, b_{hi}, b_{hf}, b_{ho}, b_{hc}$ are the bias vectors.  
$X_t$ represents the input at time $t$, while $h_{t-1}$ and $C_{t-1}$ denote the previous hidden and cell states. This hidden state is then passed to the next time step. These equations describe the flow of information through the ConvLSTM cell at each time step. The convolutional operations allow the network to capture spatial dependencies in the input data sequence, making ConvLSTM particularly effective for tasks involving sequential data with spatial characteristics, such as video processing and activity recognition \cite{Ref35,Ref36}. To train this model, it uses parameters list in Table \ref{fig:convlstm_param} and features extracted from image sequences that are illustrated.

\begin{table}[htbp]
\centering
% \tiny
\caption{ConvLSTM Model Parameters}
\label{fig:convlstm_param}
\begin{tabular}{lc}
\hline
\textbf{Parameter} & \textbf{Value} \\ \hline
Total Parameters & 266,240 \\
Trainable Parameters & 266,240 \\
Non-trainable Parameters & 0 \\
Batch Size & 16 \\
Epochs & 150 \\
Loss Function & Sparse Categorical Crossentropy \\
Optimizers & Adam, RMSprop, SGD, Adagrad \\
Activation Functions & Softmax, ReLU, Tanh \\
Learning Rate & 0.0001 \\ \hline
\end{tabular}
\end{table}

By combining spatial feature extraction with temporal sequence modeling in a single architecture, the ConvLSTM proves highly effective at capturing complex human activities, even in short, noisy video sequences.

\subsubsection{BiLSTM (Bidirectional Long Short-Term Memory)}
\begin{figure}[ht]
    \centering
    \includegraphics[width=1.0\linewidth]{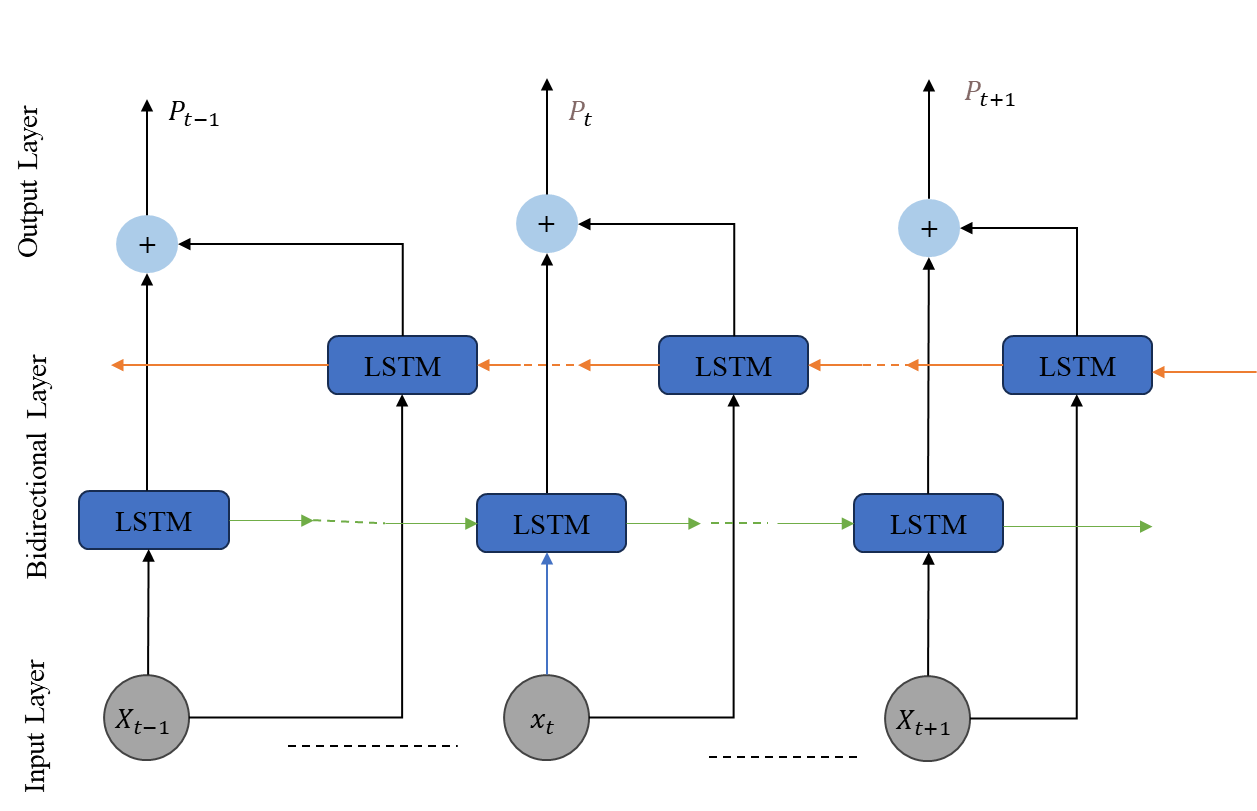} % <-- Replace image file name
    \caption{The internal architecture of BiLSTM}
    \label{fig:bilstm_arch}
\end{figure}

We adopted the specialized LSTM variant known as Deep BiLSTM, an extension of the traditional Bidirectional LSTM (BiLSTM) architecture. This model is particularly effective in analyzing sequential spatial features within the local environment, enabling the capture of long-term dependencies in spatiotemporal patterns. Figure~\ref{fig:bilstm_arch} presents the detailed architecture of the proposed Deep BiLSTM model.The corresponding model parameters are summarized in Table ~\ref{tab:bilstm_params} and classify human actions based on features extracted from image sequence. BiLSTM processes frame sequences bidirectionally, enabling richer temporal context modeling. It’s especially useful in tasks like video-based action recognition and gesture detection, where context from the whole sequence matters and regular LSTMs that only look forward, BiLSTM captures information from both before and after each moment, making it better at spotting patterns across time. Both the learnt information by above hidden states is combined in latter layer. Equation (6) to (8) explains the operations performed in BiLSTM unit. 
\begin{equation}
P_t = \sigma(W_1 x_t + W_2 P_{t-1}) \odot \tanh(C_t)
\label{eq:bilstm_forward}
\end{equation}

\begin{equation}
P_{t+1} = \sigma(W_3 x_t + W_4 P_{t-1}) \odot \tanh(C^{'}_t)
\label{eq:bilstm_backward}
\end{equation}

\begin{equation}
og_t = W_5 P_t + W_6 P_{t+1}
\label{eq:bilstm_output}
\end{equation}

where $x_t$ is the input at time $t$, $w_1, w_2, \ldots, w_6$ are the weights of gates of LSTM cells. $P_t$ are the forward and $P_{t+1}$ backward output, $og_t$ is combined output at time $t$, and $C_t$, $C_t'$ are the forward and backward LSTM respectively \cite{Ref37}. It describes how a BiLSTM computes its outputs by processing the input sequence in both forward and backward directions. At each time step $t$, the forward LSTM output $P_t$ is calculated by applying a $\sigma$ sigmoid function to a weighted sum of the current input $x_t$ and the previous output $P_{t-1}$, followed by element wise multiplication with the transformed cell state $\tanh(C_t)$. Similarly, the backward LSTM output $P_{t+1}$ is computed similarly but processes the sequence in reverse, using separate weights and its own cell state $C_t'$. Finally, the outputs from both directions are combined using a weighted sum to produce the final BiLSTM output $og_t$ at time $t$. This fusion of forward and backward context allows the BiLSTM to effectively capture video information from both past and future time steps, improving its performance on sequential tasks in HAR \cite{Ref38}. As a result, BiLSTM and its deeper variants remain widely adopted in the development of intelligent, context aware HAR systems.
\begin{table}[htbp]
\centering
% \tiny
\caption{BiLSTM Model Parameters}
\label{tab:bilstm_params}
\begin{tabular}{lc}
\hline
\textbf{Parameter} & \textbf{Value} \\ \hline
Total Parameters & 8,138,104 \\
Trainable Parameters & 8,095,640 \\
Non-trainable Parameters & 42,464 \\
Batch Size & 16 \\
Epochs & 150 \\
Loss Function & Sparse Categorical Crossentropy \\
Optimizers & Adam, RMSprop, SGD, Adagrad \\
Activation Functions & Softmax, ReLU, Tanh \\
Learning Rate & 0.0001 \\ \hline
\end{tabular}
\end{table}

We experiment with multiple AF within the hidden layers and MO to evaluate their influence on convergence behavior and classification accuracy. This configuration allows the network to capture not only what is happening in each frame but also how actions evolve over time, resulting in a robust model for complex HAR in short, noisy video sequences. All experiments used batch size = 16. BiLSTM was trained for 150 epochs using a learning rate of 1e-3. Loss function: sparse categorical cross-entropy. Data augmentation: random crop, horizontal flip, brightness jitter. BiLSTM remains a widely adopted model in intelligent and context-aware HAR systems due to its high accuracy and robust temporal modeling ability.
\begin{table*}[t]
\centering
% \tiny
\caption{Comparison of commonly used activation functions in deep learning-based HAR models.}
\label{tab:af_comparison_twocolumn}
\renewcommand{\arraystretch}{1.35}
\setlength{\tabcolsep}{6pt}
\begin{tabular}{|p{2.6cm}|p{3.0cm}|p{4.8cm}|p{4.8cm}|}
\hline
\textbf{Activation Function} & \textbf{Equation} & \textbf{Advantages} & \textbf{Drawbacks} \\ \hline

ReLU~\cite{Ref39} & 
$f(x)=\max(0,x)$ & 
Efficient for HAR video tasks due to fast convergence; effectively captures prominent spatial features. & 
Can cause dead neurons; ignores negative activations, potentially limiting the representation of fine temporal patterns. \\ \hline

Sigmoid~\cite{Ref40} & 
$f(x)=\frac{1}{1+e^{-x}}$ & 
Produces smooth probabilistic outputs; suitable for binary or confidence-based HAR decisions. & 
Suffers from severe vanishing gradient issues in deeper architectures, reducing training efficiency. \\ \hline

Tanh~\cite{Ref41} & 
$f(x)=\frac{e^x - e^{-x}}{e^x + e^{-x}}$ & 
Zero-centered output improves sequential modeling; often preferred for LSTM-based HAR tasks. & 
Slower training compared to ReLU; still vulnerable to gradient saturation in deep networks. \\ \hline

\end{tabular}
\end{table*}

\begin{table*}[ht]
\centering
% \tiny % Use a smaller font size
\caption{Model Optimizer Comparison}
\begin{tabular}{|p{2cm}|p{2.6cm}|p{4cm}|p{4cm}|} % Adjusted column widths
\hline
\textbf{Optimizer} & \textbf{Equation} & \textbf{Advantages} & \textbf{Drawbacks} \\ \hline

SGD~\cite{Ref42} &
$ w_j^{(t+1)} = w_j^{(t)} - \eta \cdot \frac{\partial J(w)}{\partial w_j } $ &
Good for simple HAR tasks; learns steadily. &
Slow training, struggles with noisy data, needs tuning. \\ \hline

Adam~\cite{Ref43} &
$ w_{t+1} = w_t - \eta \cdot \frac{\hat{m}_t}{\sqrt{\hat{v}_t} + \epsilon} $ &
Fast for complex tasks, handles large datasets. &
Uses more memory, risk of overfitting, struggles with sparse data. \\ \hline

RMSProp~\cite{Ref44} &
$ w_{t+1} = w_t - \eta \cdot \frac{g_t}{\sqrt{v_t} + \epsilon} $ &
Good for dynamic patterns, effective in LSTM models. &
Unstable for sparse data, requires careful tuning. \\ \hline

Adagrad~\cite{Ref45} &
$ w_{t+1} = w_t - \eta \cdot \frac{g_t}{\sqrt{G_t} + \epsilon} $ &
Good for sparse data, adapts learning rate. &
Slows down over time, struggles with dense data, may stop learning early. \\ \hline

\end{tabular}
\label{tab:optimizer_comparison_twocolumn}
\end{table*}
\subsection{Activation Functions (AF)}

AFs introduce non-linearity into deep learning models, enabling them to learn complex patterns and representations from sensor or video data.Without activation functions, deep networks would behave like a simple linear regression model, limiting their capability to handle real-world data intricacies found in HAR tasks. The choice of activation function can affect the Model convergence speed, Representation power, Gradient flow during backpropagation, Overall accuracy and generalization capability. This research considers only three activation functions named ReLU (Rectified Linear Unit), Sigmoid and Tanh (Hyperbolic Tangent). All of the Activation functions have some major characteristics and make them different from each other’s.

\subsubsection{ReLU}
ReLU (Rectified Linear Unit) is a computationally efficient activation function that helps reduce the vanishing gradient problem by allowing gradients to propagate through the network without saturating. It is particularly effective in CNN-based and hybrid HAR models, where speed and performance are critical. ReLU also encourages sparse activation, meaning fewer neurons are activated at once, leading to better generalization and improved model efficiency.
\subsubsection{Sigmoid}
Sigmoid is commonly used in binary classification tasks due to its output range of 0 to 1, which is interpretable as probability. It is often applied in the output layer of models to provide a clear decision boundary. However, Sigmoid suffers from the vanishing gradient problem, especially in deep networks, which can slow down training and hinder the model's ability to learn effectively.

\subsubsection{Tanh}
Tanh outputs values between -1 and 1, providing a more balanced activation range compared to Sigmoid, making it ideal for sequence modeling tasks like those in RNNs and LSTMs. Despite its advantages in balancing activations, Tanh is slower to compute and still suffers from the vanishing gradient problem, especially in deep networks, limiting its use in very deep architectures. These behaviours are particularly relevant in HAR tasks, where models must balance fast convergence with stable gradient flow. A concise comparison of the three activation functions used in this work is provided in Table~\ref{tab:af_comparison_twocolumn}.
We focus on AF and MO variations because prior HAR studies rarely analyze their joint impact, despite strong influence on gradient stability and convergence in recurrent video models.

\subsection{Model Optimizers (MO)}
MO are used to update the weights of the neural network to minimize the loss function. The efficiency and stability of the training process in HAR models significantly depend on the choice of the optimizer. Optimizer selection influences the training convergence speed, stability of learning, ability to escape local minima and the performance on imbalanced or noisy HAR datasets. This research considers only four MOs and Each model is trained using all combinations of AF and MO. All of the MO has some major characteristics and make them different from each other’s.

\subsubsection{Stochastic Gradient Descent (SGD)}
Stochastic Gradient Descent (SGD) is a simple and widely used optimization algorithm in machine learning. It updates the model weights by calculating the gradient of the loss function with respect to a single data point, rather than the entire dataset. This makes it computationally efficient and suitable for large datasets. However, SGD requires careful tuning of the learning rate, and it can be slower compared to more advanced optimizers, especially when the learning rate is not optimal.

\subsubsection{Adaptive Moment Estimation (Adam)}
Adaptive Moment Estimation (Adam) is a popular optimization algorithm that combines the benefits of RMSProp and momentum. It adapts the learning rate for each parameter based on first and second moments of the gradients. Adam is known for its fast convergence and is highly effective when dealing with noisy gradients. This makes it particularly useful for deep architectures like ConvLSTM, where training can be complex and time-consuming.

\subsubsection{RMSprop}
RMSprop (Root Mean Square Propagation) is an optimization algorithm designed to address the issues of non-stationary targets in online and non-convex problems. It works by adapting the learning rate for each parameter, scaling it based on the moving average of squared gradients. This property makes RMSprop particularly well-suited for Recurrent Neural Networks (RNNs) and Human Activity Recognition (HAR) tasks, where data distributions can change over time.

\subsubsection{Adagrad}
Adagrad is an adaptive learning rate algorithm that adjusts the learning rate based on the historical gradients of each parameter. This ensures that parameters with frequent updates have a smaller learning rate, while those with infrequent updates have a larger learning rate. Adagrad is particularly effective for sparse data, such as sensor data in HAR tasks, where some features may be more relevant than others and need a different learning rate.

Although the optimizers described in Table \ref{tab:optimizer_comparison_twocolumn} share the common objective of reducing the loss function, each follows a distinct update strategy that directly influences the training behavior of deep HAR models. SGD provides stable gradient updates but typically requires manual tuning to achieve fast convergence. Adam and RMSProp offer adaptive learning rates, making them effective for complex temporal models where gradient magnitudes vary significantly across frames. Adagrad performs well when dealing with sparse or irregular input patterns but may experience diminishing learning rates during long training cycles. By applying all four optimizers across both BiLSTM and ConvLSTM architectures, this study demonstrates how differences in update rules, adaptability, and convergence characteristics shape model performance on HAR datasets. These observations help clarify which optimizers are most suitable for reliable and efficient training in activity-recognition applications, particularly those used in healthcare and real-time monitoring environments.

\section{Result Analysis}
In this research, ConvLSTM and BiLSTM are used as end-to-end sequence models that operate directly on video frame sequences from both datasets. Each model learns both spatial and temporal patterns from the data and outputs the final activity class predictions. Due to space limitations, only summary accuracy metrics are provided in Tables~\ref{tab:convLSTM} and \ref{tab:BiLSTM}. The complete training and validation accuracy/loss curves for all activation function–optimizer combinations are presented in Table~\ref{tab:traincurves}, which groups all plots in a consolidated format.
% ============================================================
%             RESULT ANALYSIS – FIGURES IN TABLE FORMAT
% ============================================================
\onecolumn

\begin{center}
\begin{longtable}{c p{0.45\linewidth} p{0.45\linewidth}}
\caption{Performance Comparison of ConvLSTM and BiLSTM Models (Accuracy and Loss Plots)} \\
\label{tab:traincurves} \\
\hline
& \textbf{Accuracy Plot} & \textbf{Loss Plot} \\ \hline
\endfirsthead

% ====================== 1. ConvLSTM – ReLU ======================
&
\begin{minipage}{\linewidth}
\centering
\includegraphics[width=\linewidth]{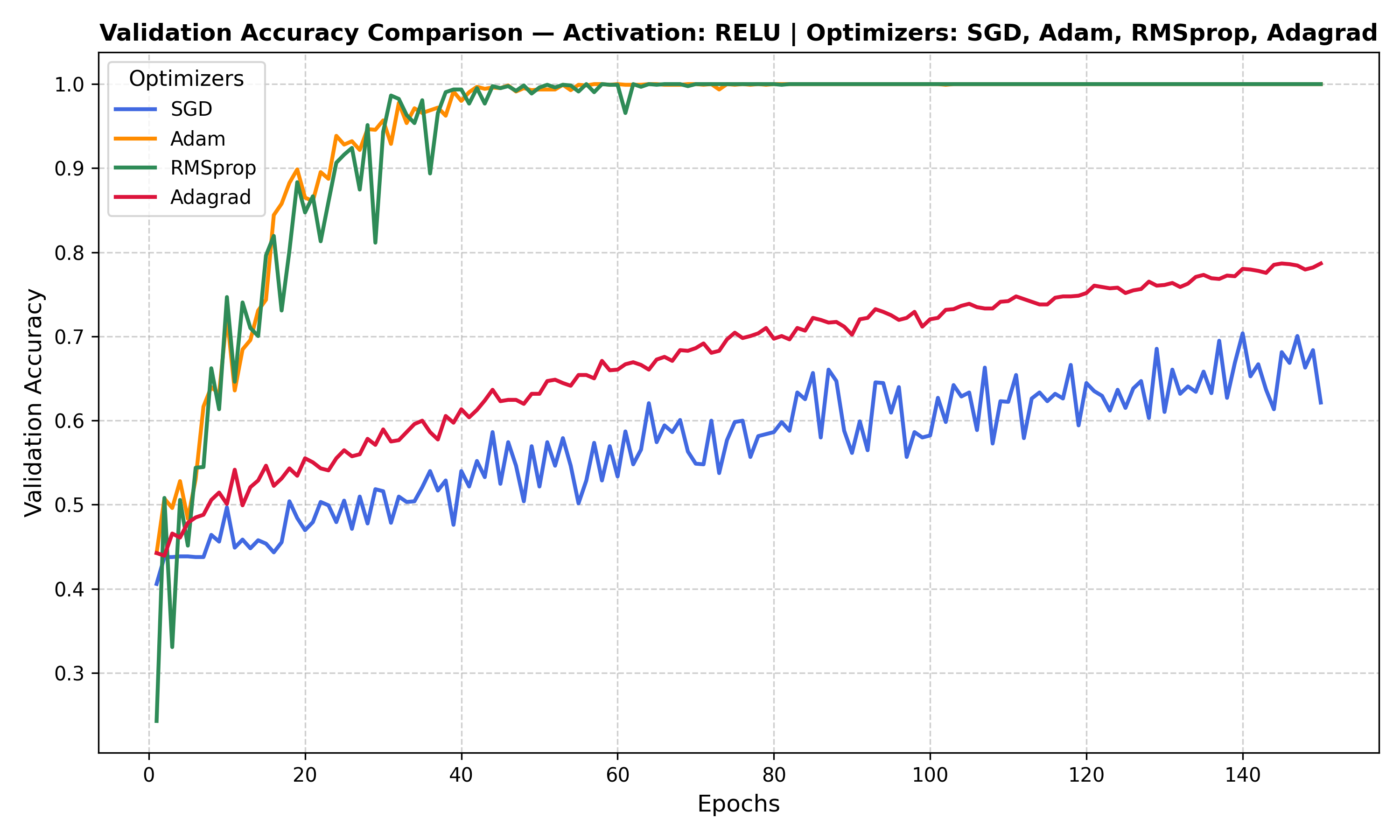}
\textbf{Figure 6(a).} Accuracy on HMDB51 using ReLU
\end{minipage}
&
\begin{minipage}{\linewidth}
\centering
\includegraphics[width=\linewidth]{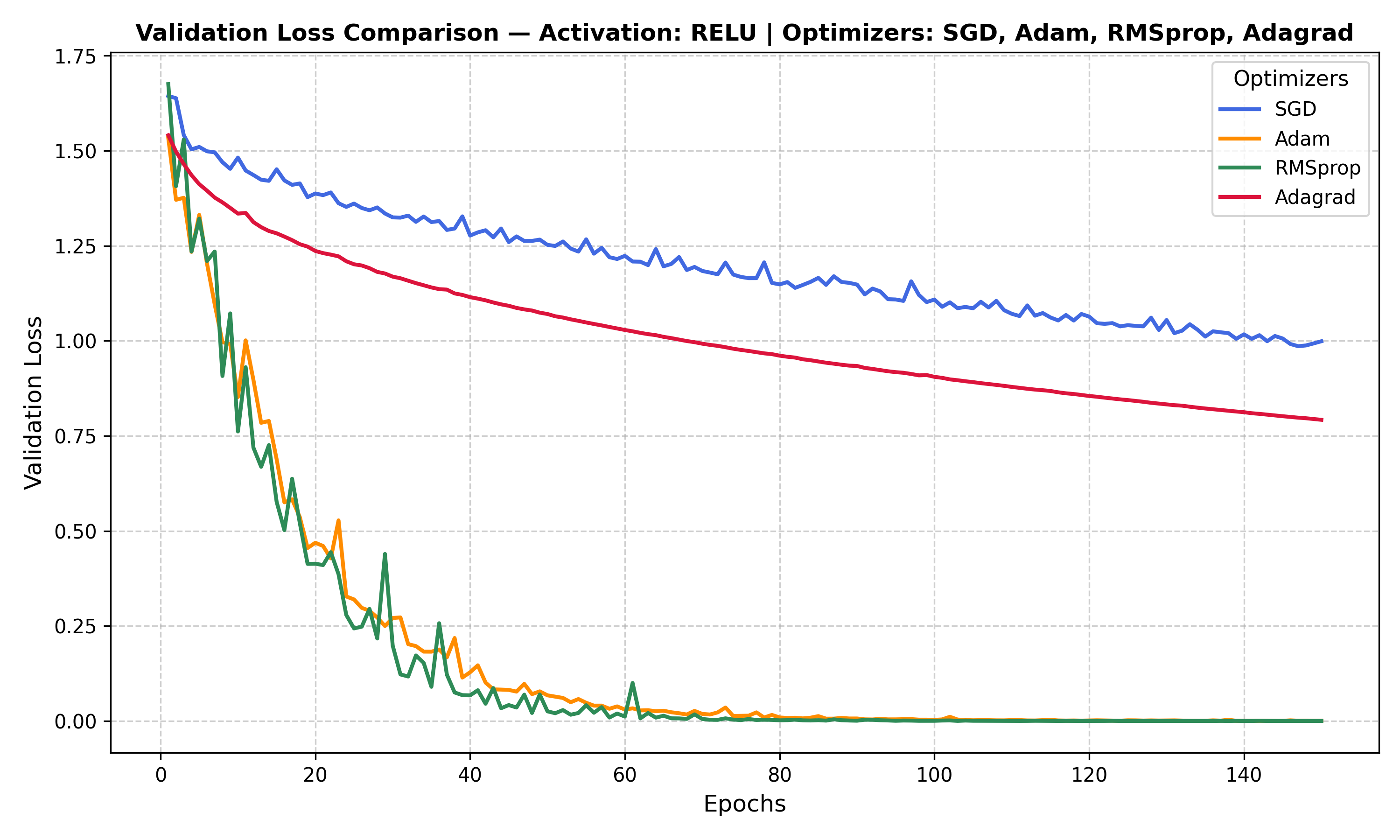}
\textbf{Figure 6(b).} Loss on HMDB51 using ReLU
\end{minipage}
\\ \hline

&
\begin{minipage}{\linewidth}
\centering
\includegraphics[width=\linewidth]{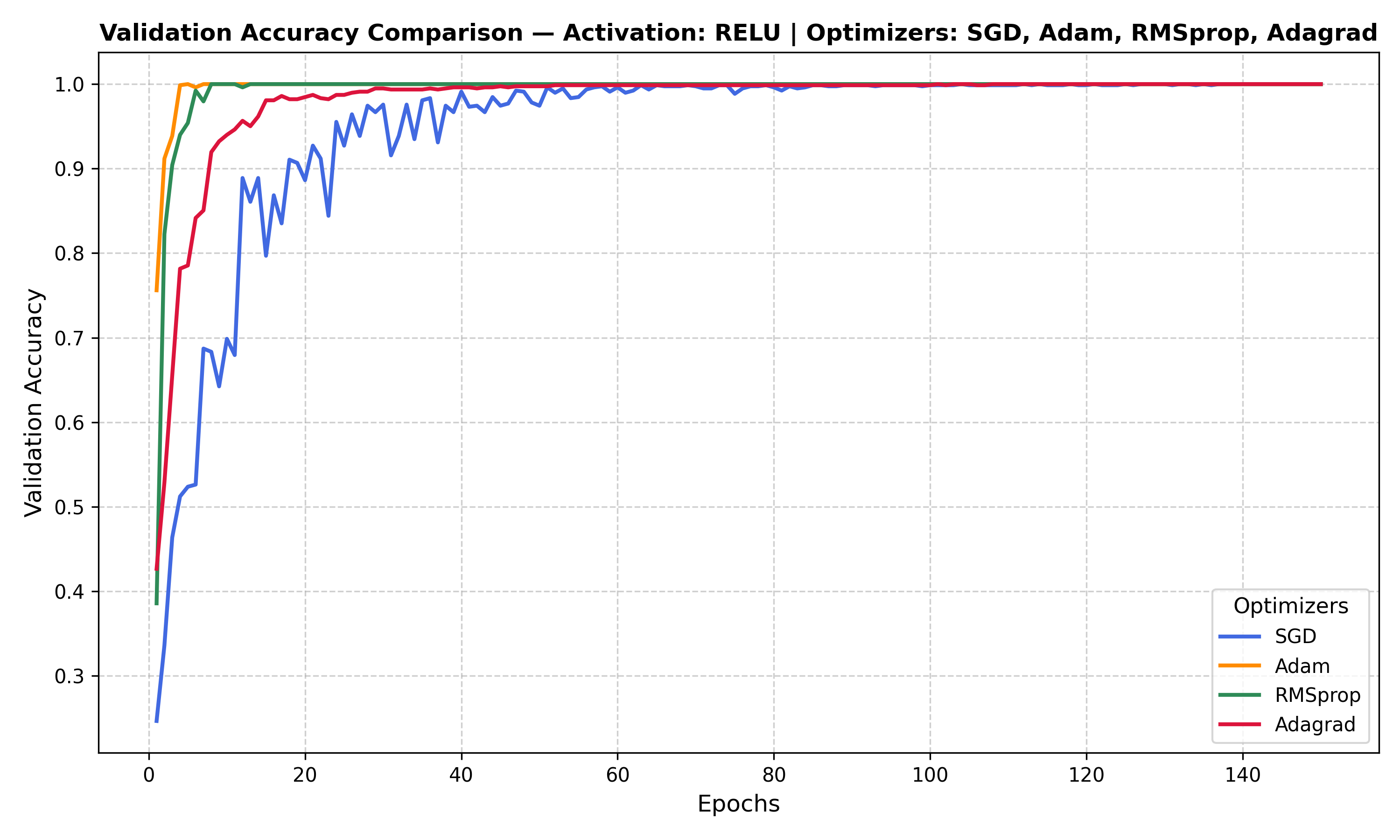}
\textbf{Figure 6(c).} Accuracy on UCF101 using ReLU 
\end{minipage}
&
\begin{minipage}{\linewidth}
\centering
\includegraphics[width=\linewidth]{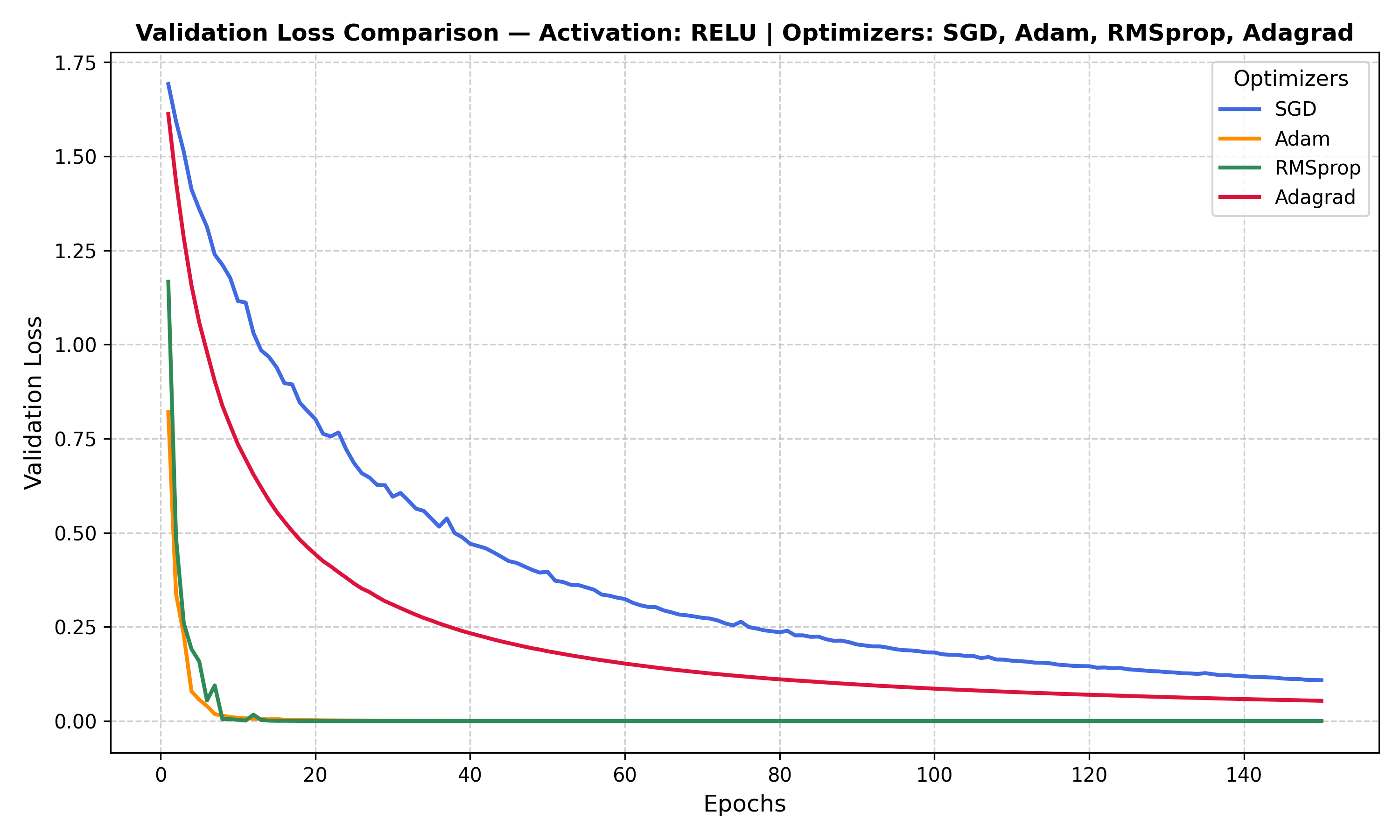}
\textbf{Figure 6(d).} Loss on UCF101 using ReLU
\end{minipage}
\\ \hline

% ====================== 2. ConvLSTM – Sigmoid ======================

&
\begin{minipage}{\linewidth}
\centering
\includegraphics[width=\linewidth]{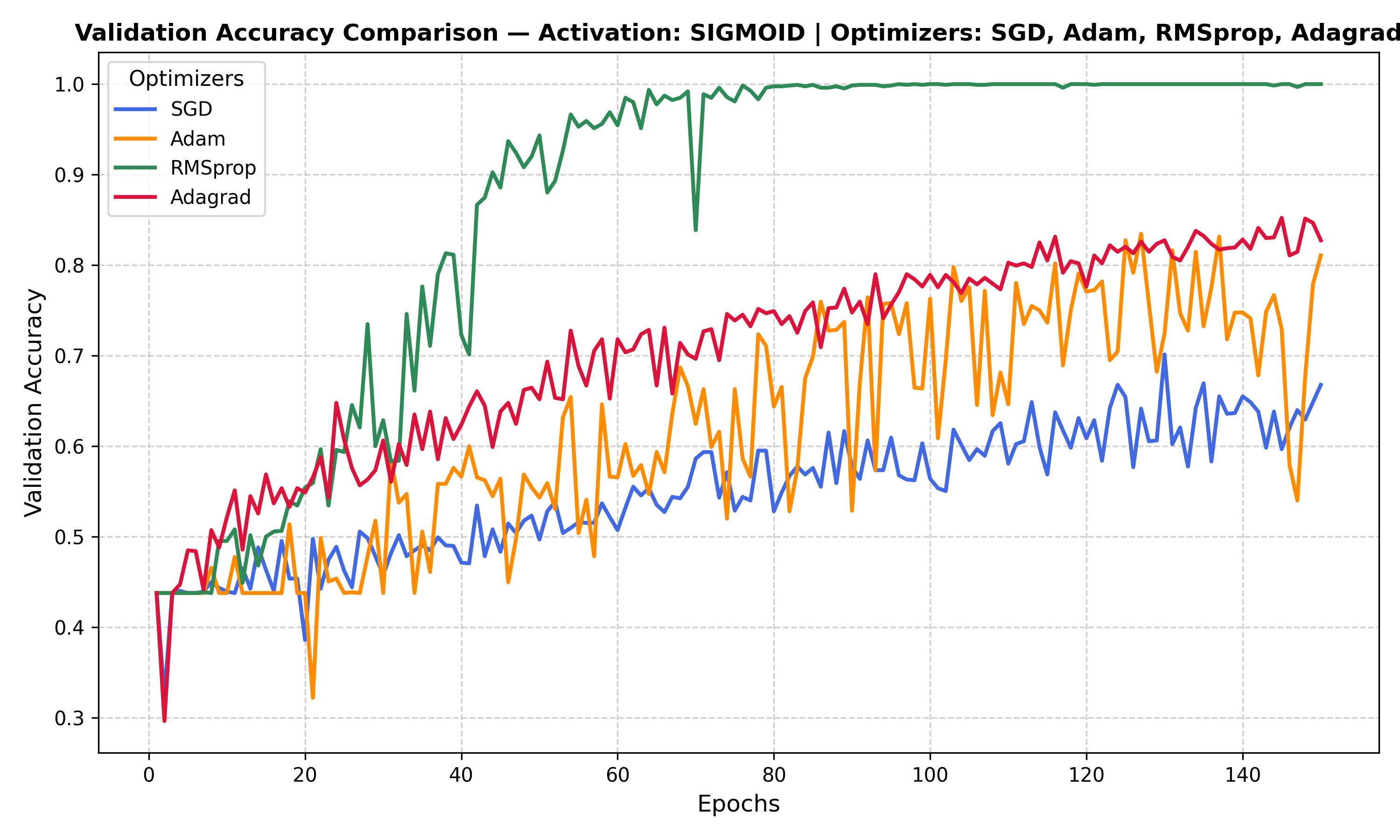}
\textbf{Figure 7(a).} Accuracy on HMDB51 using Sigmoid 
\end{minipage}
&
\begin{minipage}{\linewidth}
\centering
\includegraphics[width=\linewidth]{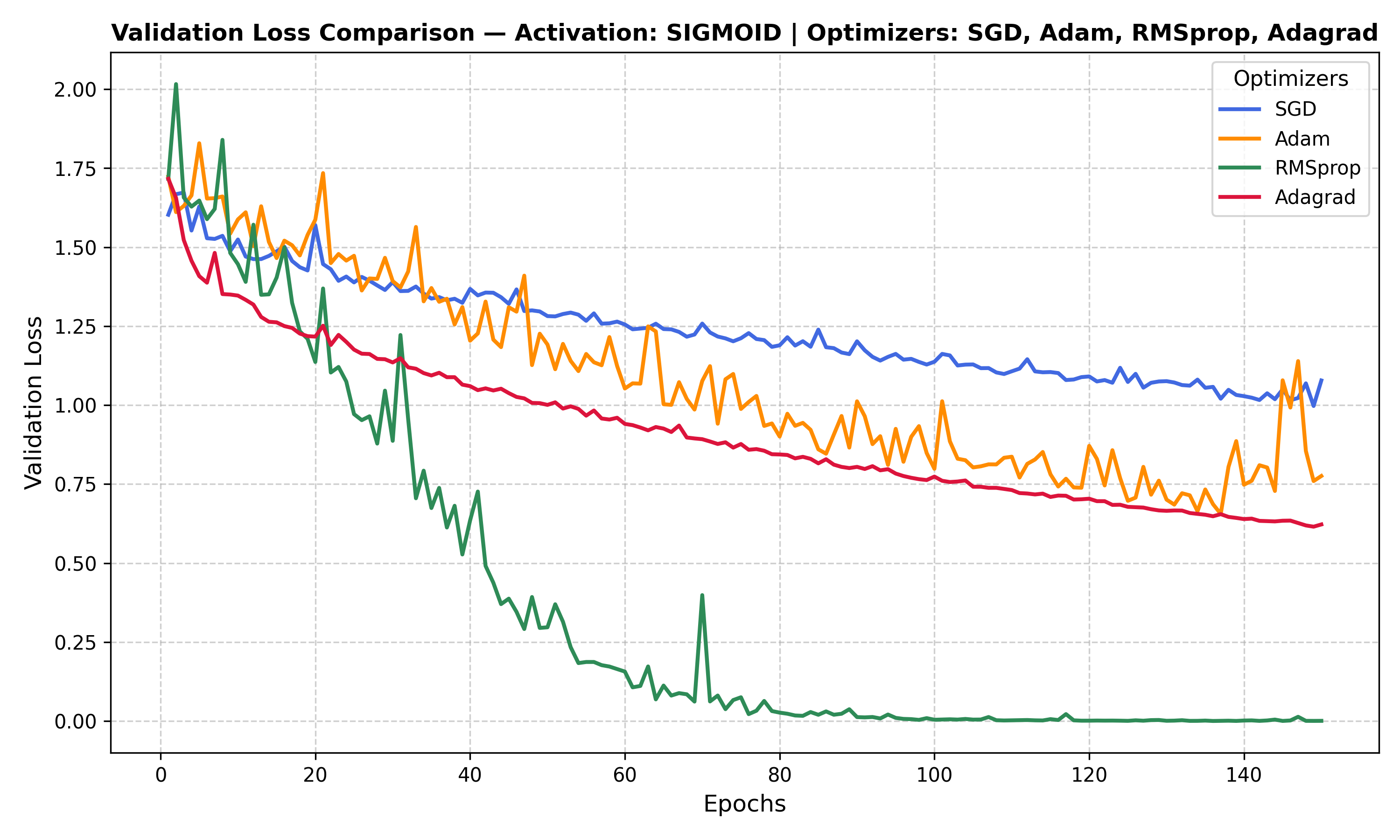}
\textbf{Figure 7(b).} Loss on HMDB51 using Sigmoid 
\end{minipage}
\\ \hline

&
\begin{minipage}{\linewidth}
\centering
\includegraphics[width=\linewidth]{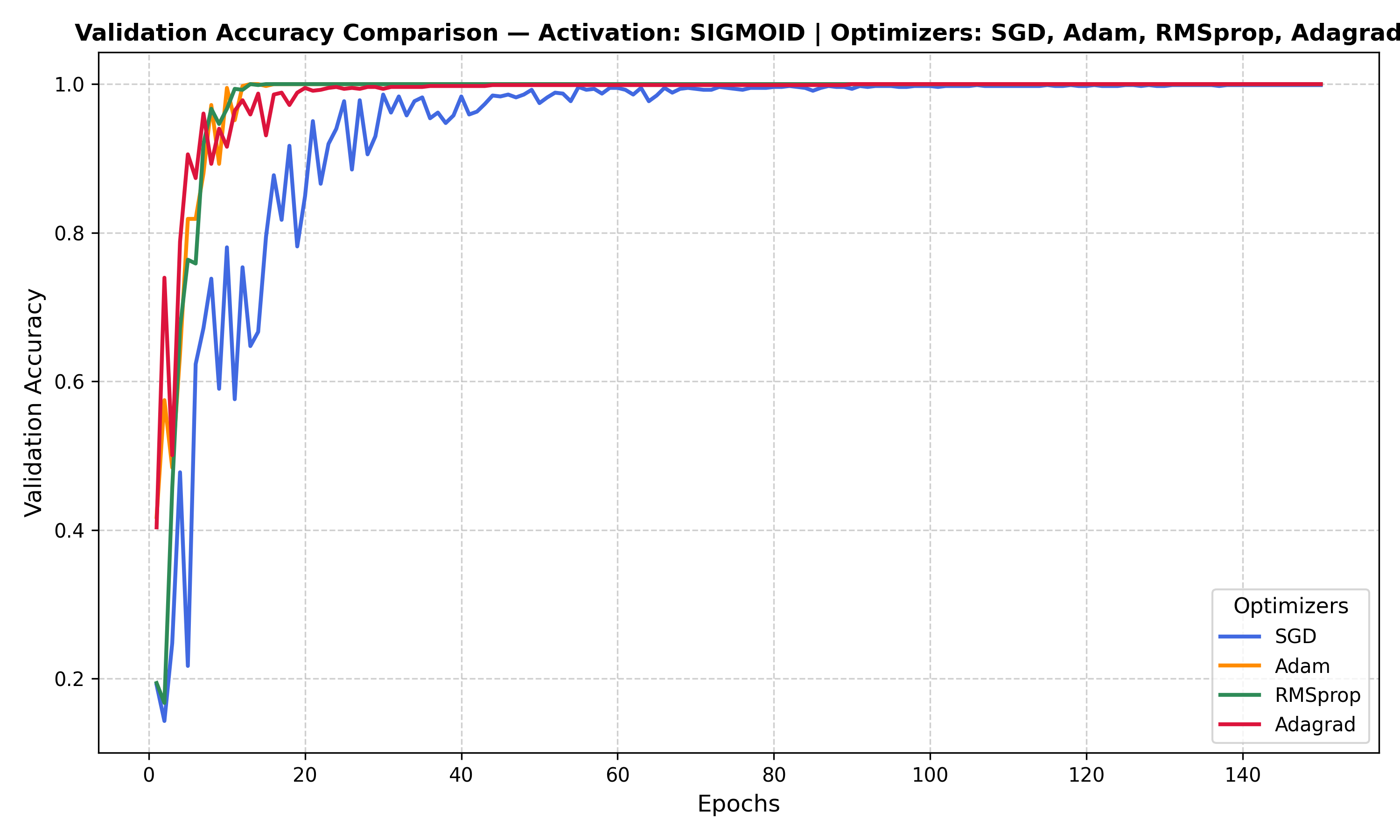}
\textbf{Figure 7(c).} Accuracy on UCF101 using Sigmoid 
\end{minipage}
&
\begin{minipage}{\linewidth}
\centering
\includegraphics[width=\linewidth]{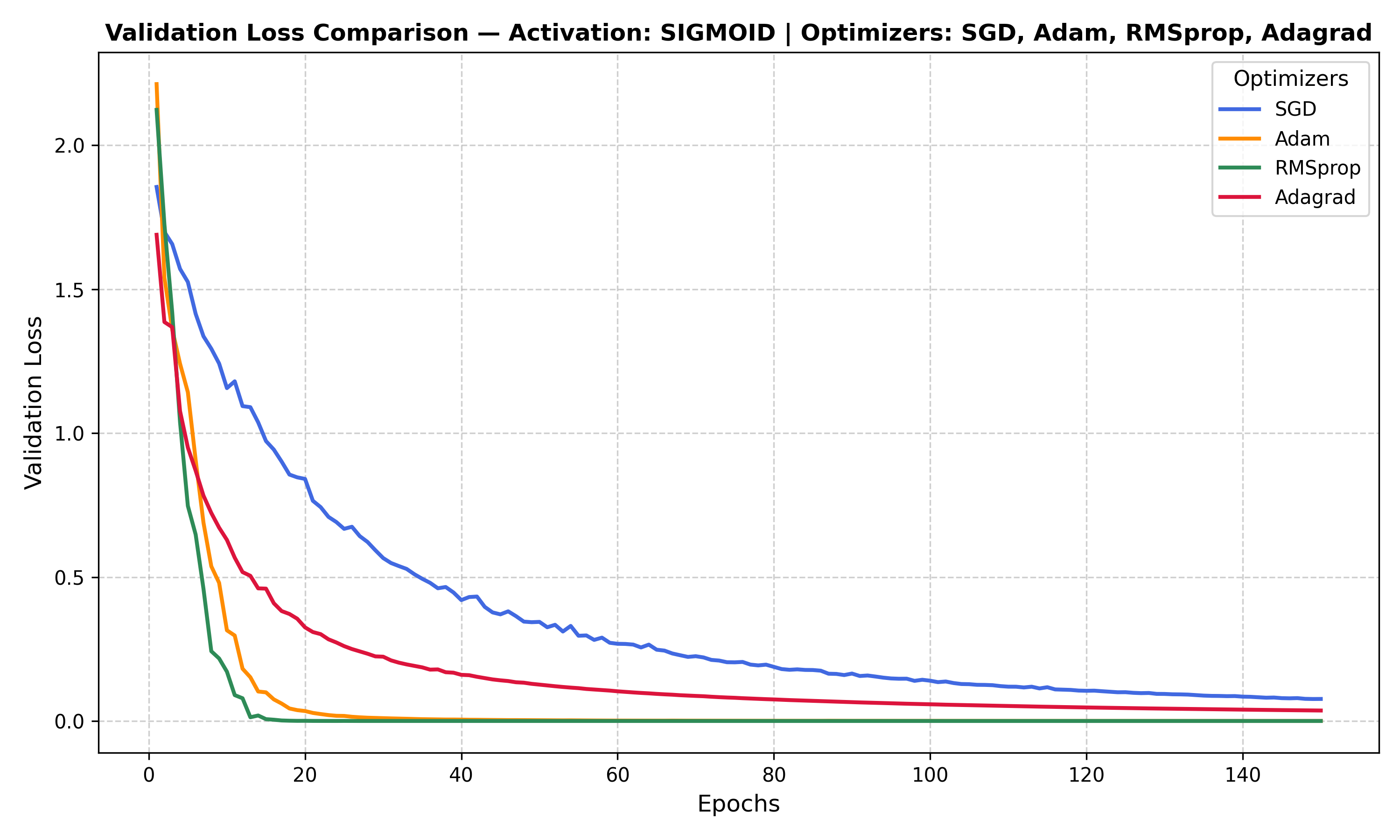}
\textbf{Figure 7(d).} Loss on UCF101 using Sigmoid 
\end{minipage}
\\ \hline

% ====================== 3. ConvLSTM – Tanh ======================
 
 &
\begin{minipage}{\linewidth}
\centering
\includegraphics[width=\linewidth]{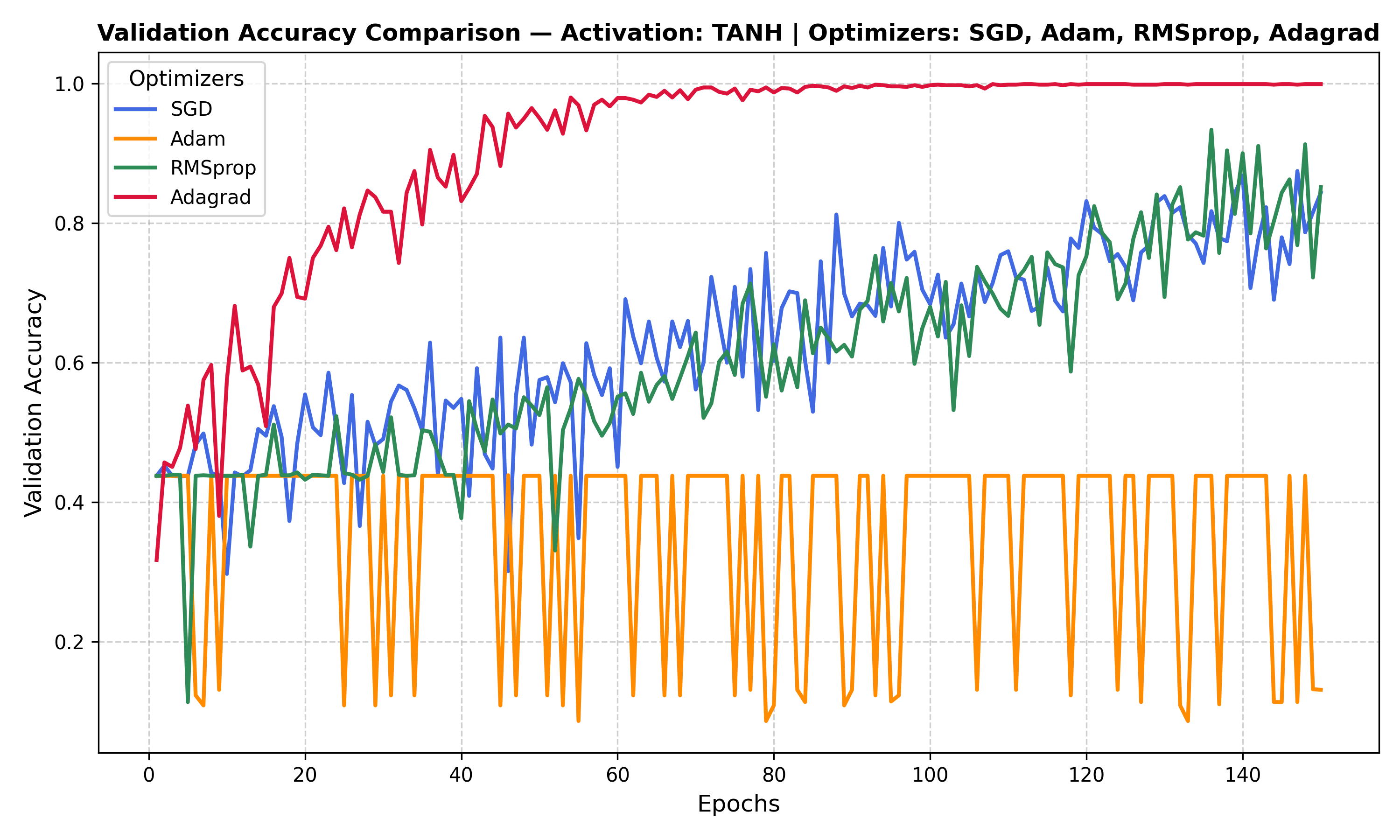}
\textbf{Figure 8(a).} Accuracy on HMDB51 using Tanh
\end{minipage}
&
\begin{minipage}{\linewidth}
\centering
\includegraphics[width=\linewidth]{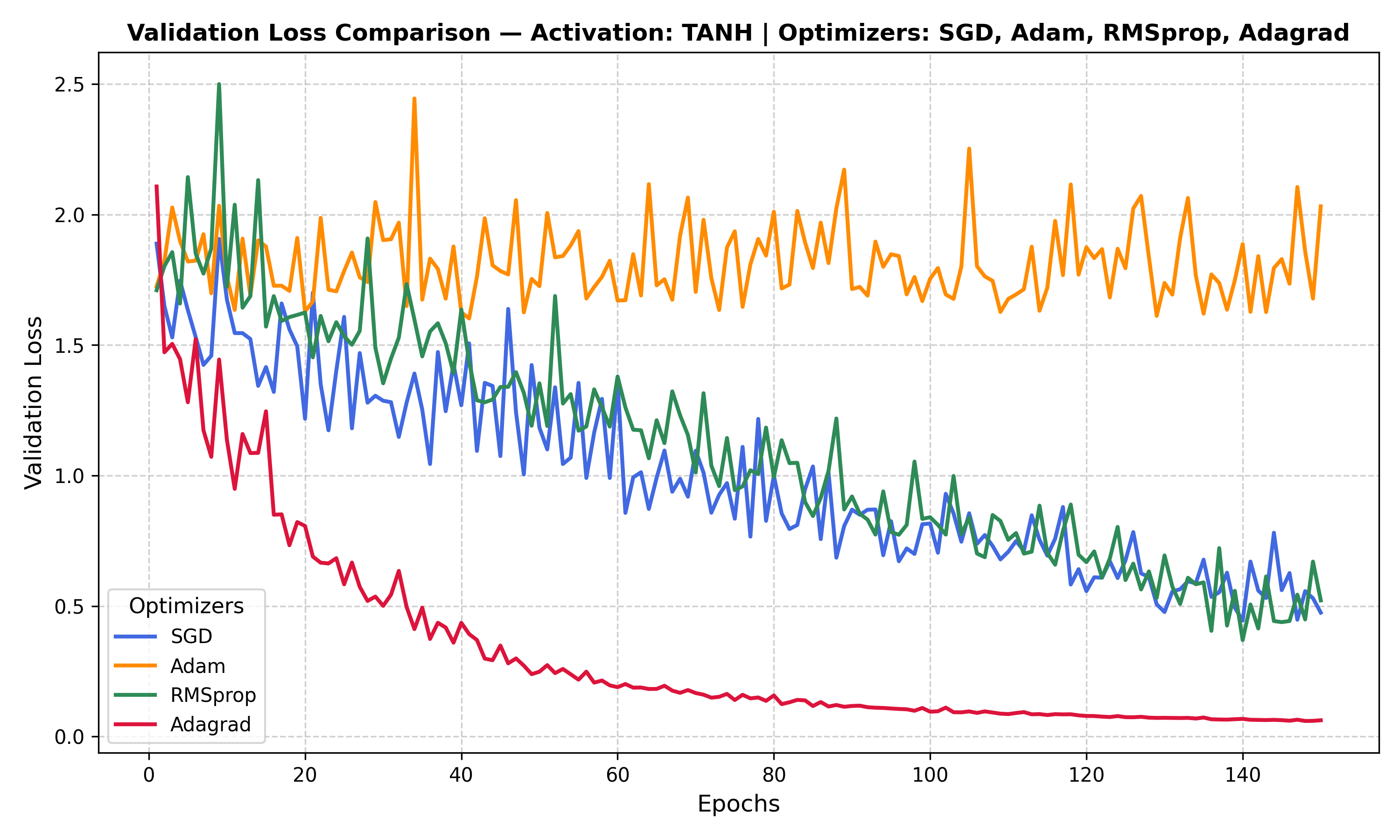}
\textbf{Figure 8(b).} Loss on HMDB51 using Tanh 
\end{minipage}
\\ \hline

&
\begin{minipage}{\linewidth}
\centering
\includegraphics[width=\linewidth]{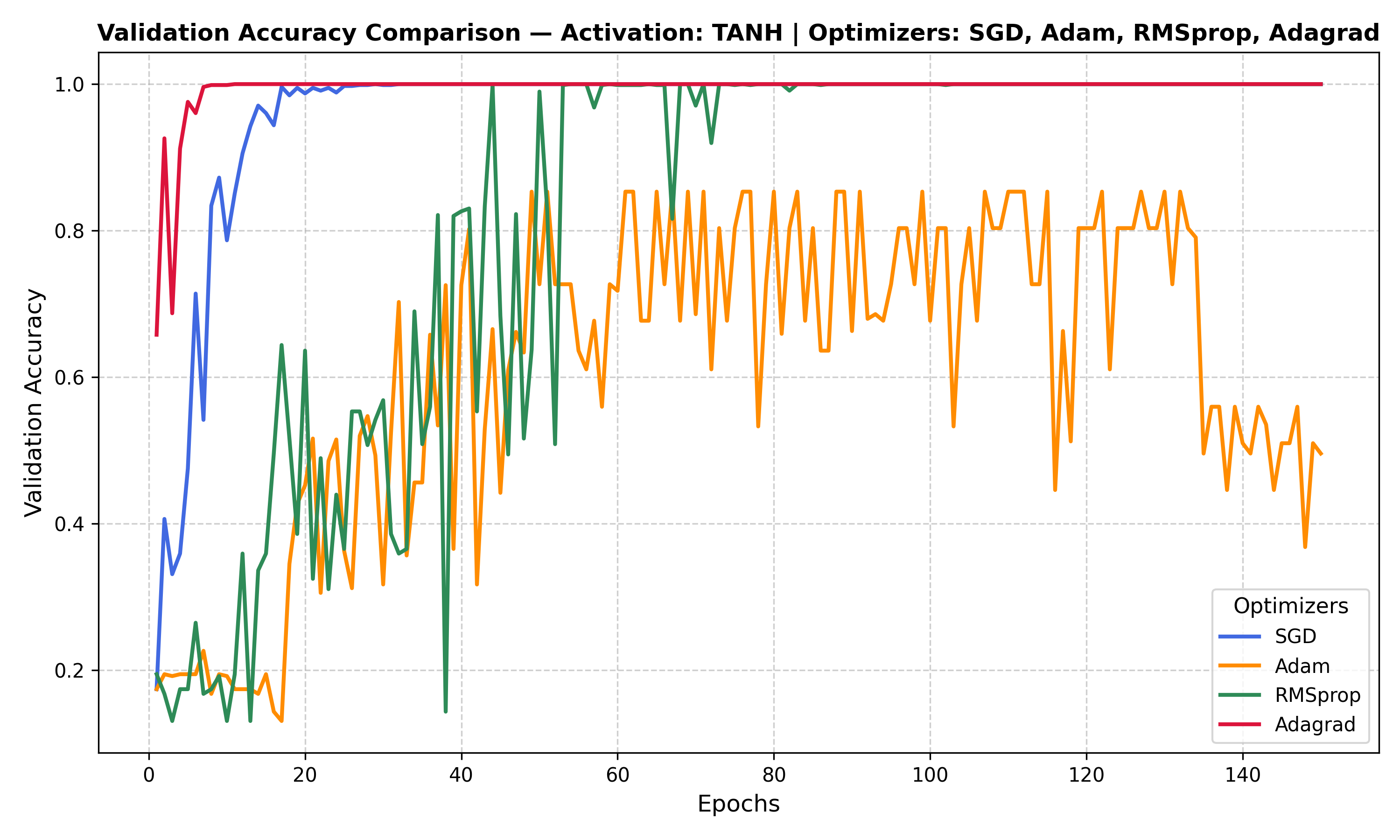}
\textbf{Figure 8(c).} Accuracy on UCF101 using Tanh 
\end{minipage}
&
\begin{minipage}{\linewidth}
\centering
\includegraphics[width=\linewidth]{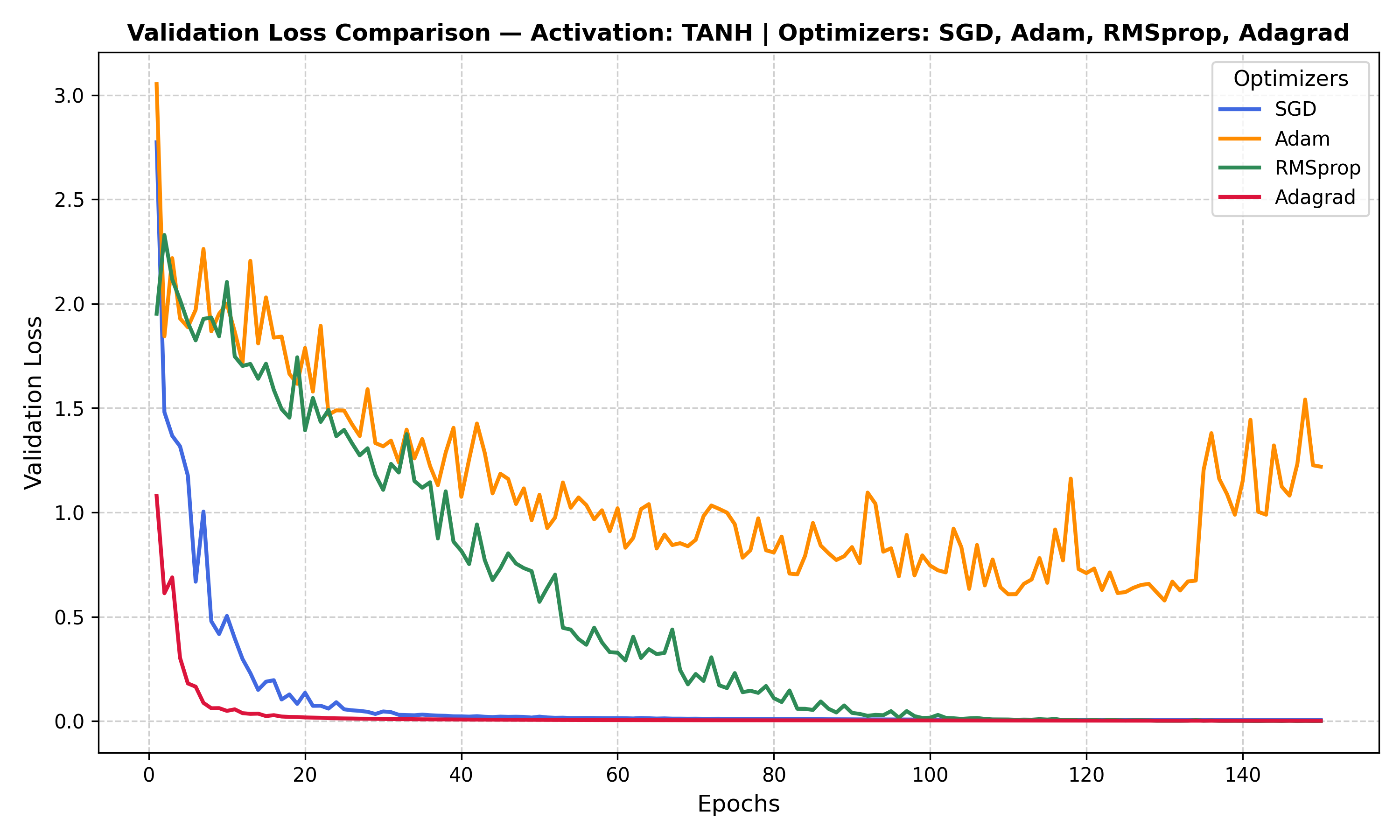}
\textbf{Figure 8(d).} Loss on UCF101 using Tanh
\end{minipage}
\\ \hline

% ====================== 4. BiLSTM – ReLU ======================
 
 &
\begin{minipage}{\linewidth}
\centering
\includegraphics[width=\linewidth]{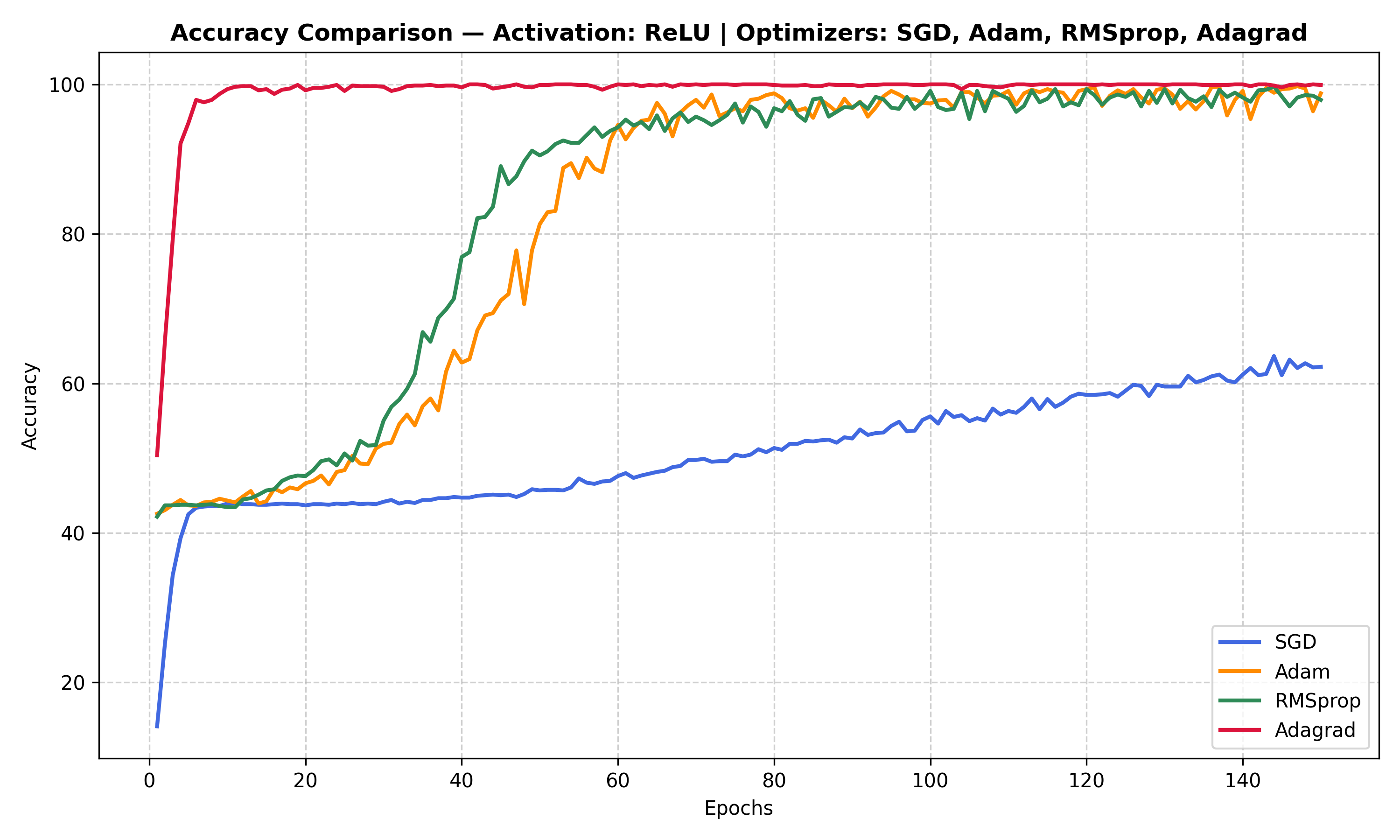}
\textbf{Figure 9(a).} Accuracy on HMDB51 using ReLU 
\end{minipage}
&
\begin{minipage}{\linewidth}
\centering
\includegraphics[width=\linewidth]{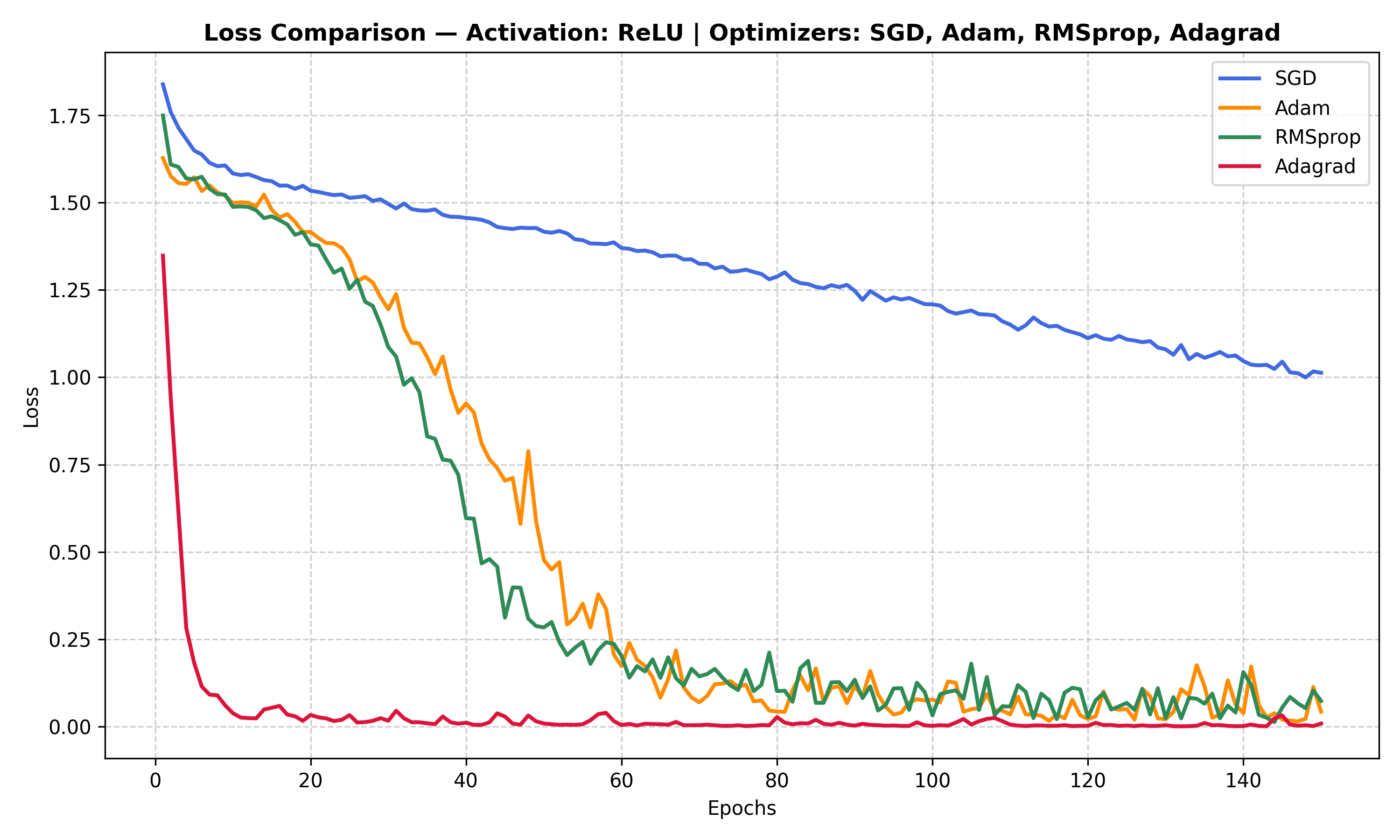}
\textbf{Figure 9(b).} Loss on HMDB51 using ReLU 
\end{minipage}
\\ \hline

 &
\begin{minipage}{\linewidth}
\centering
\includegraphics[width=\linewidth]{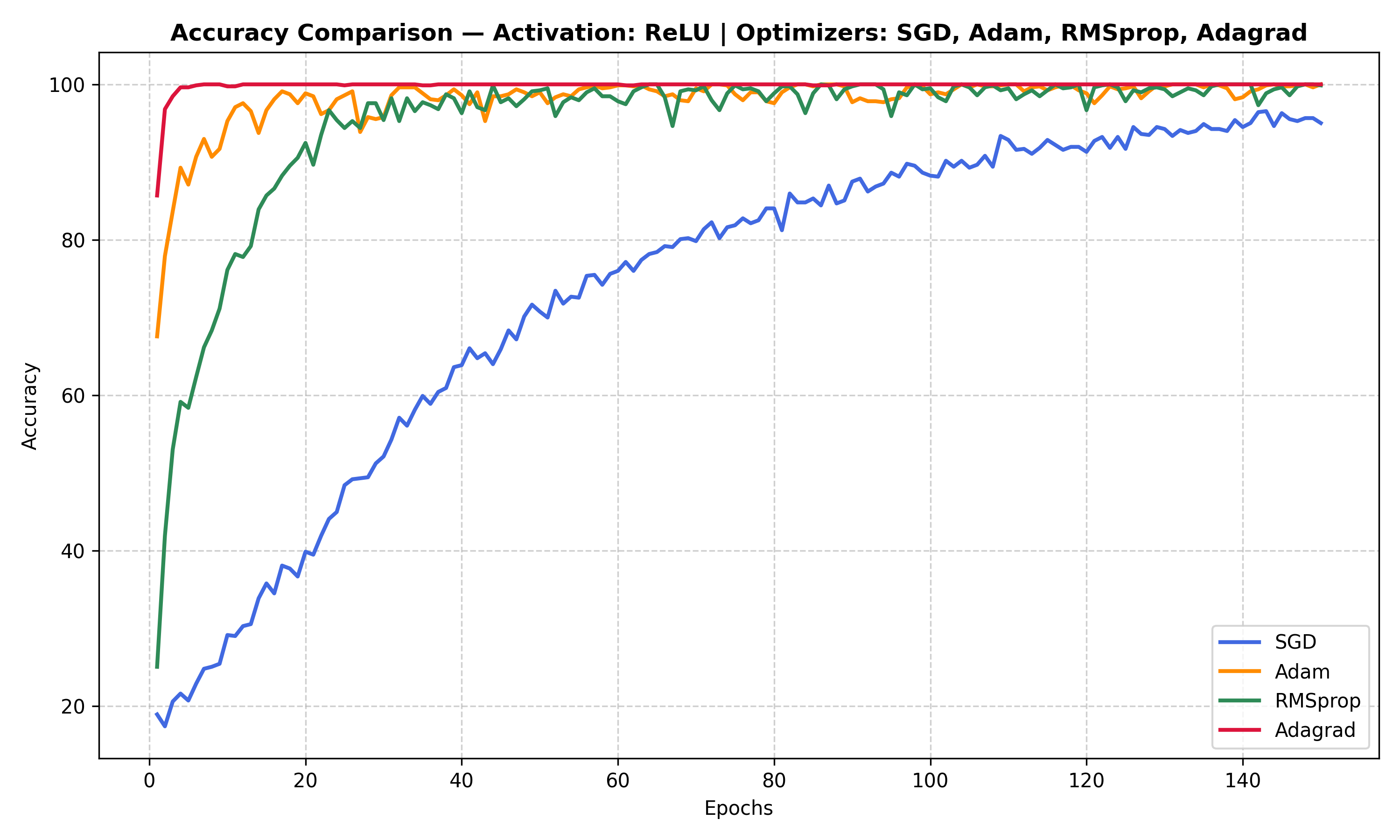}
\textbf{Figure 9(c).} Accuracy on UCF101 using ReLU 
\end{minipage}
&
\begin{minipage}{\linewidth}
\centering
\includegraphics[width=\linewidth]{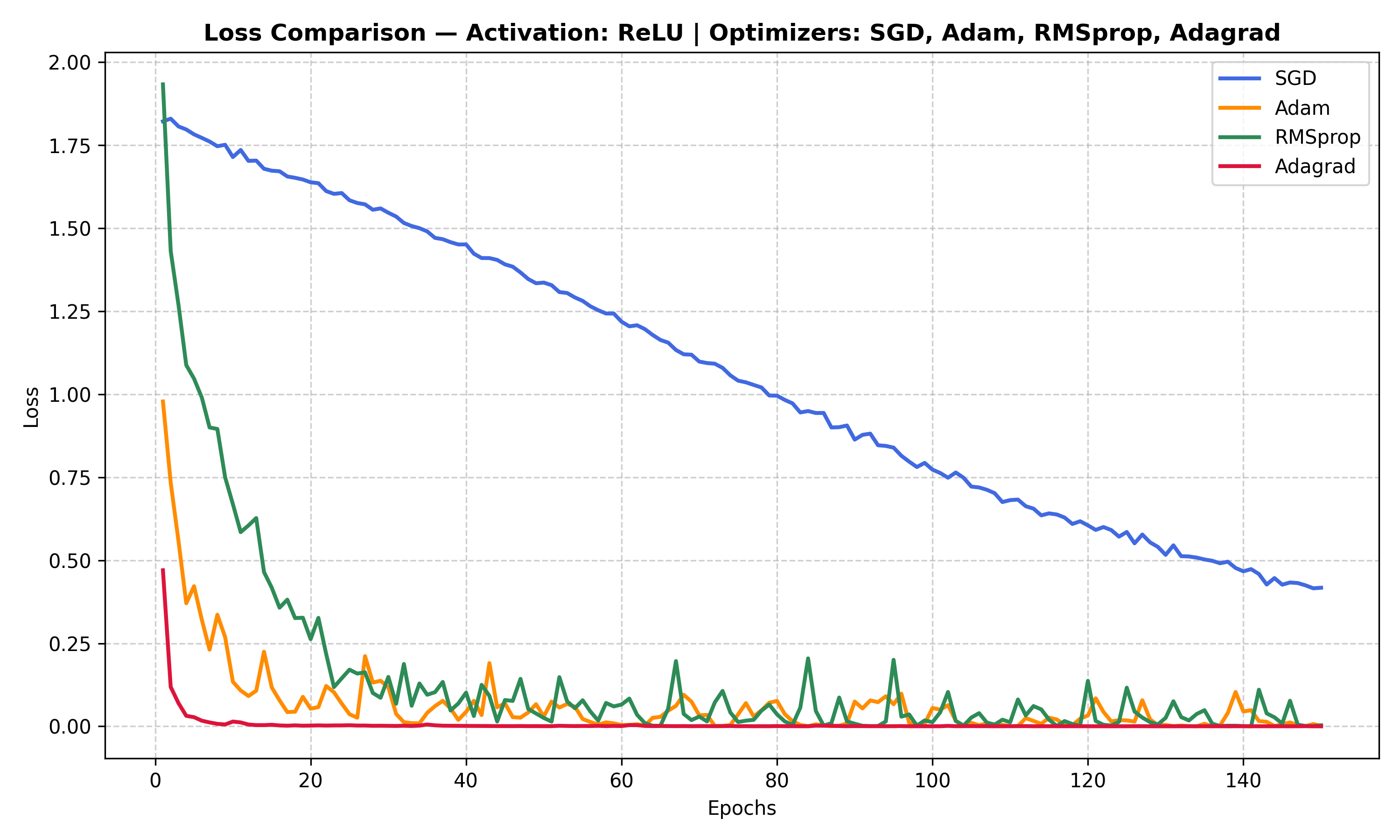}
\textbf{Figure 9(d).} Loss on UCF101 using ReLU 
\end{minipage}
\\ \hline

% ====================== 5. BiLSTM – Sigmoid ======================
 
 &
\begin{minipage}{\linewidth}
\centering
\includegraphics[width=\linewidth]{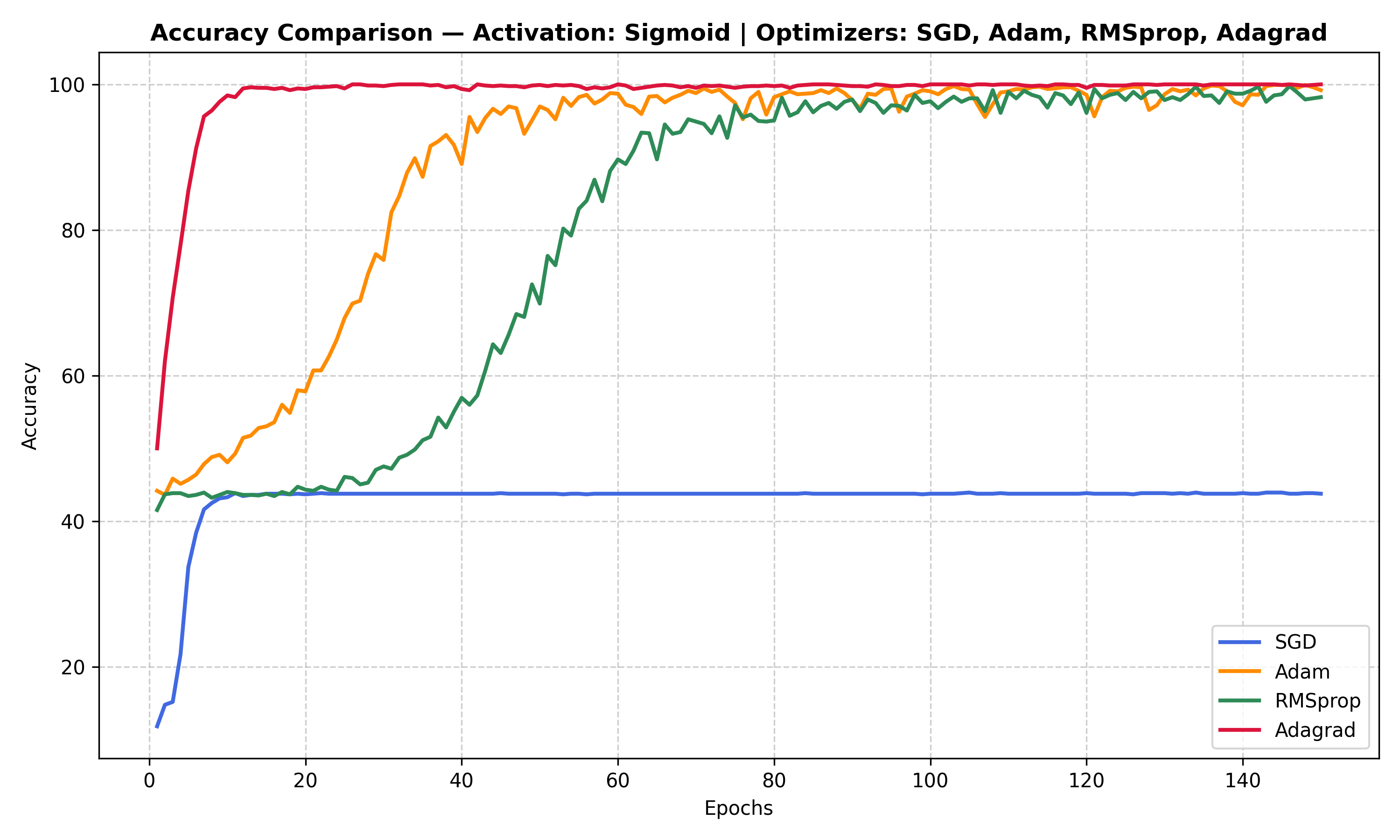}
\textbf{Figure 10(a).} Accuracy on HMDB51 using Sigmoid 
\end{minipage}
&
\begin{minipage}{\linewidth}
\centering
\includegraphics[width=\linewidth]{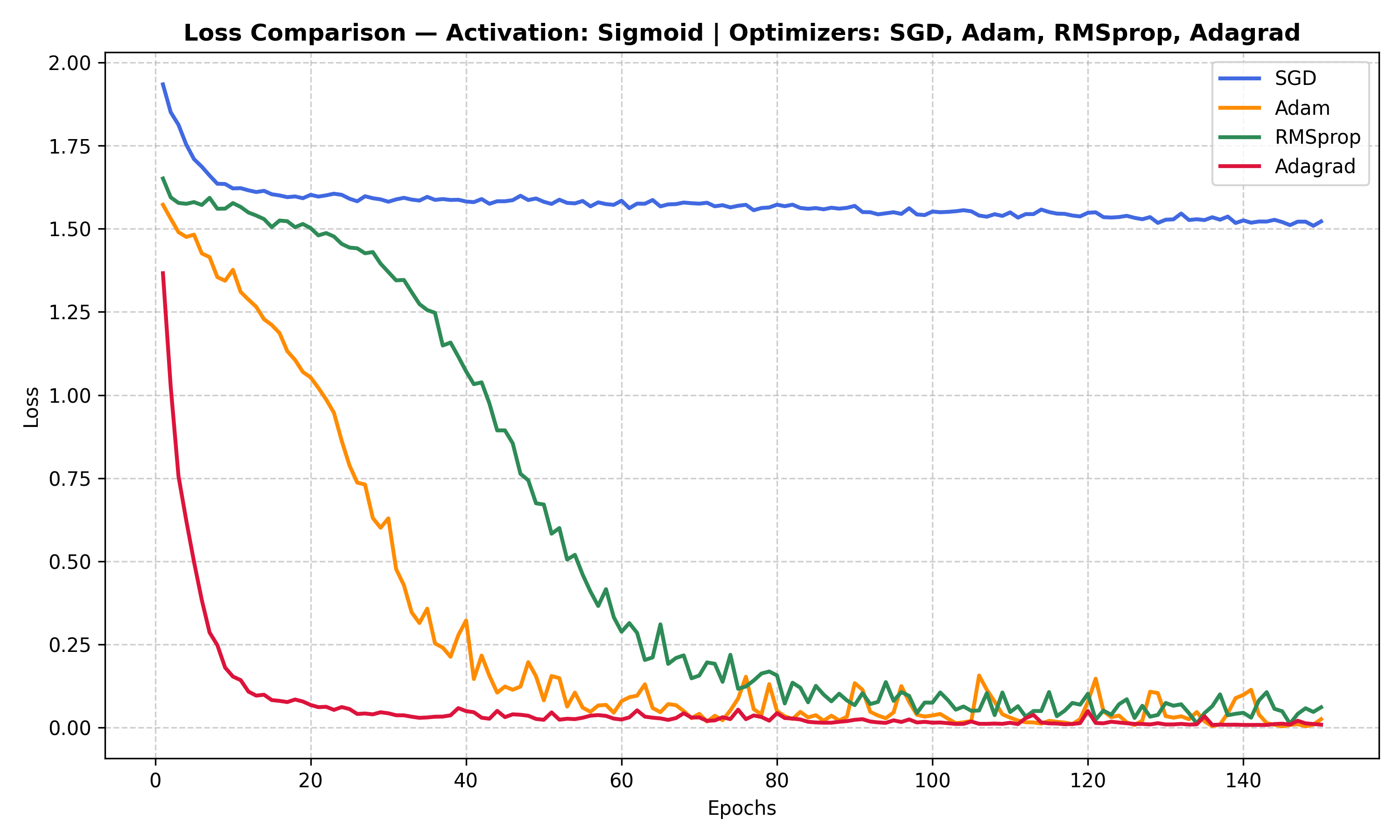}
\textbf{Figure 10(b).} Loss on HMDB51 using Sigmoid 
\end{minipage}
\\ \hline

 &
\begin{minipage}{\linewidth}
\centering
\includegraphics[width=\linewidth]{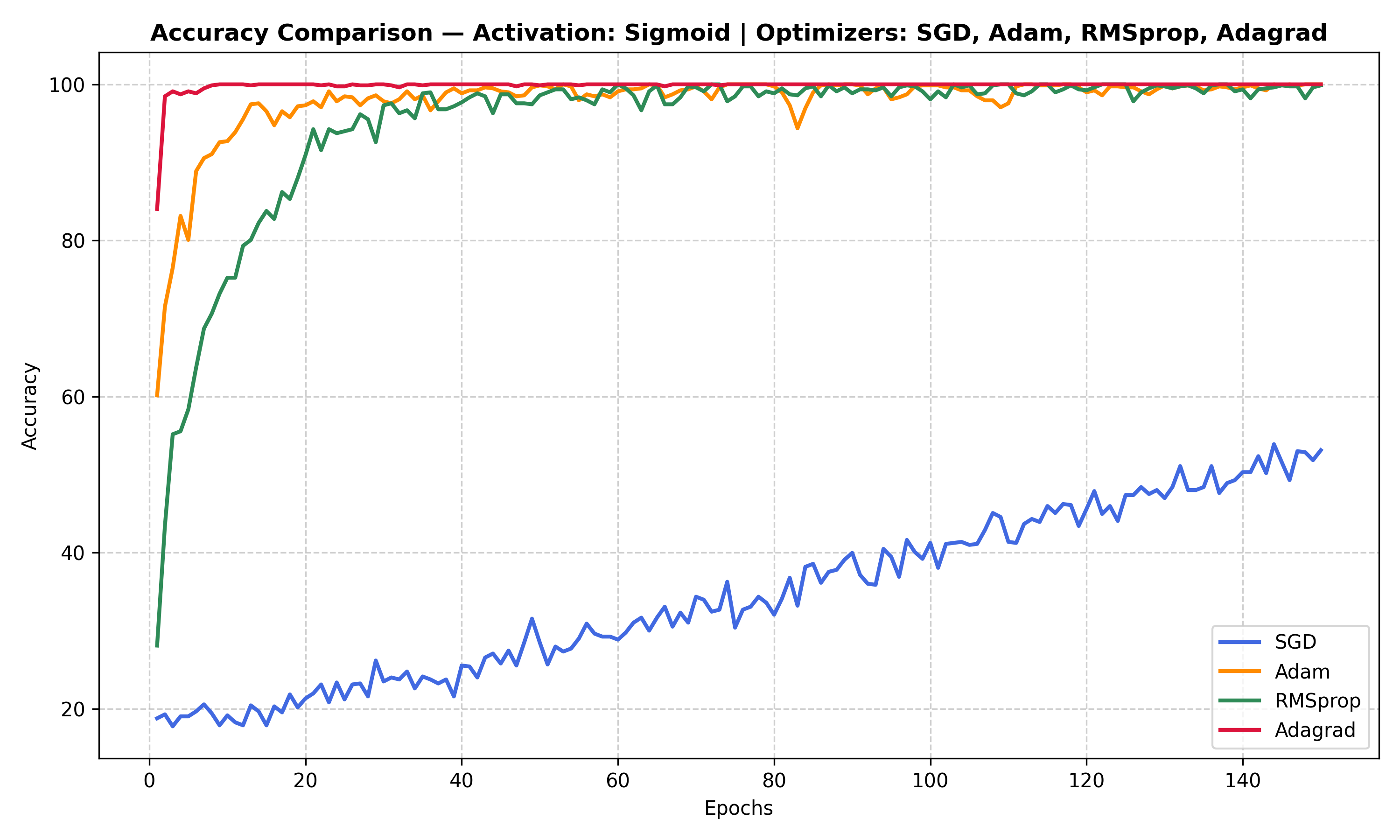}
\textbf{Figure 10(c).} Accuracy on UCF101 using Sigmoid 
\end{minipage}
&
\begin{minipage}{\linewidth}
\centering
\includegraphics[width=\linewidth]{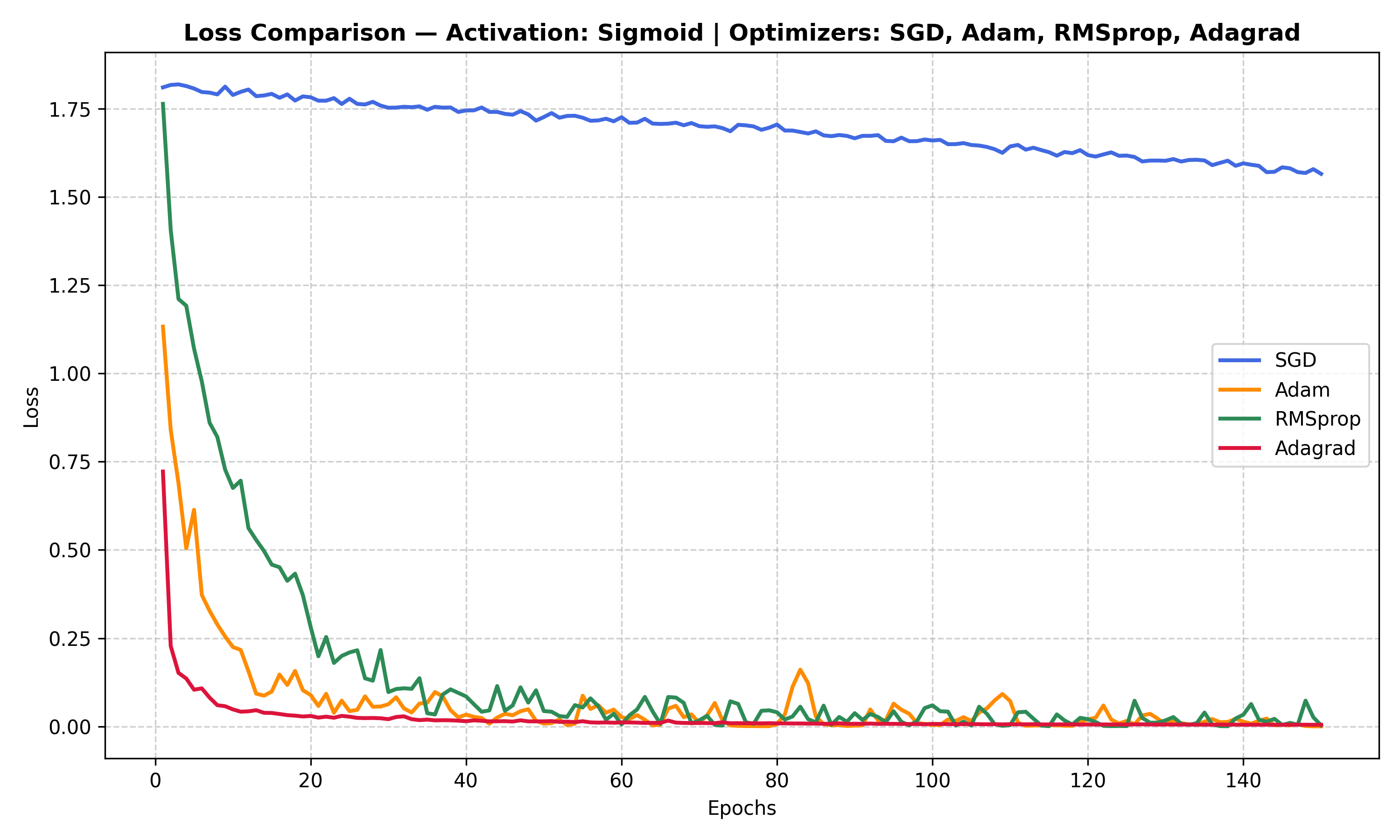}
\textbf{Figure 10(d).} Loss on UCF101 using Sigmoid 
\end{minipage}
\\ \hline

% ====================== 6. BiLSTM – Tanh ======================

&
\begin{minipage}{\linewidth}
\centering
\includegraphics[width=\linewidth]{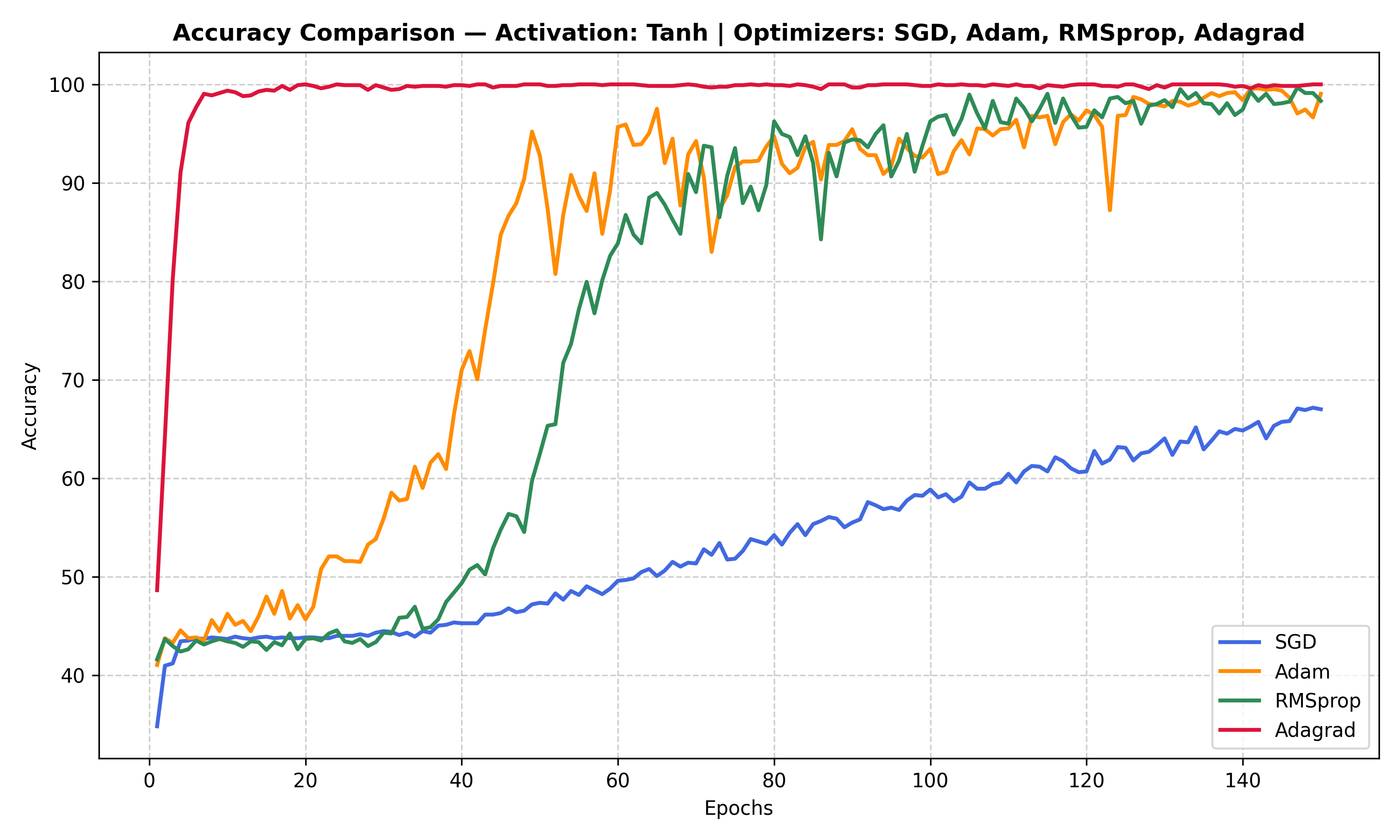}
\textbf{Figure 11(a).} Accuracy on HMDB51 using Tanh 
\end{minipage}
&
\begin{minipage}{\linewidth}
\centering
\includegraphics[width=\linewidth]{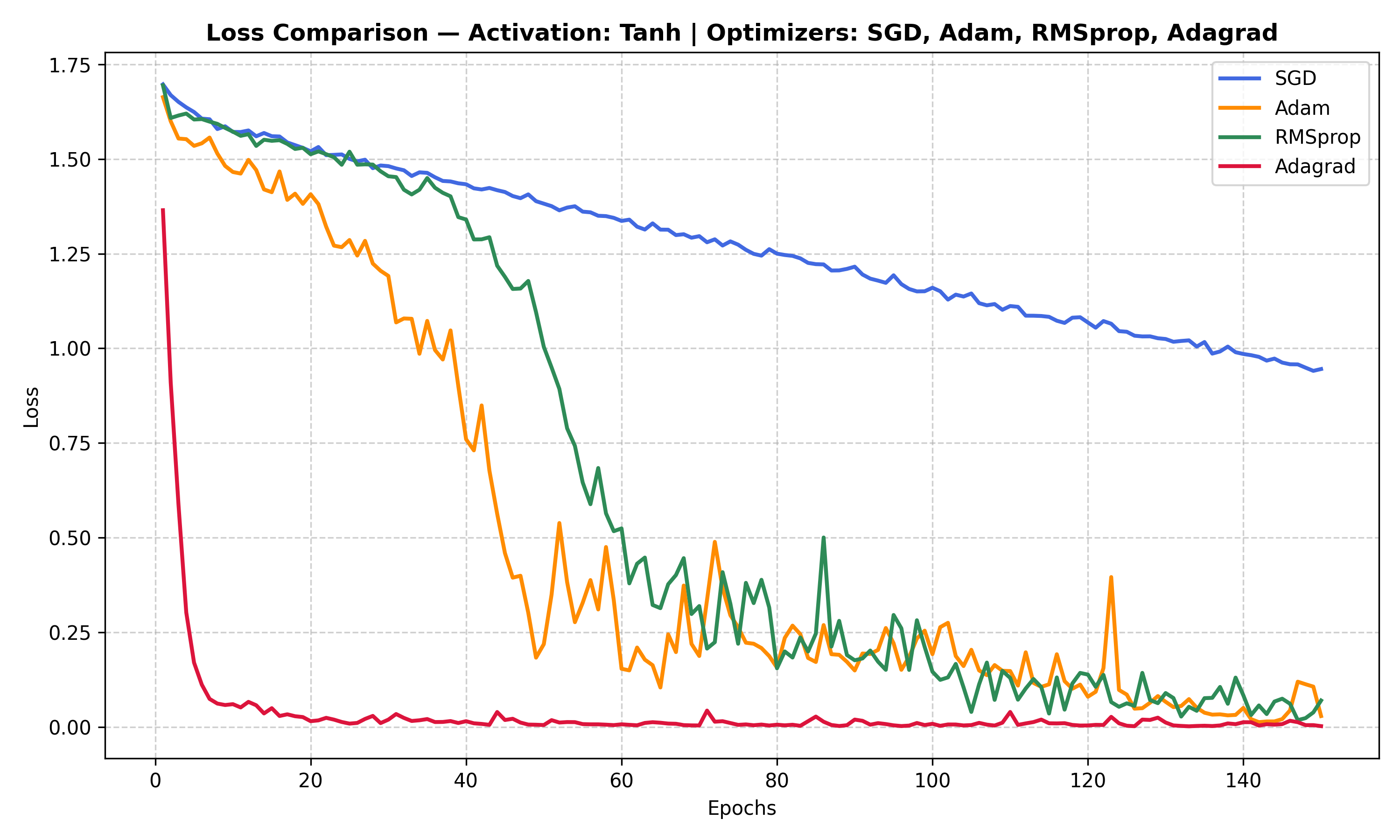}
\textbf{Figure 11(b).} Loss on HMDB51 using Tanh 
\end{minipage}
\\ \hline
 &
\begin{minipage}{\linewidth}
\centering
\includegraphics[width=\linewidth]{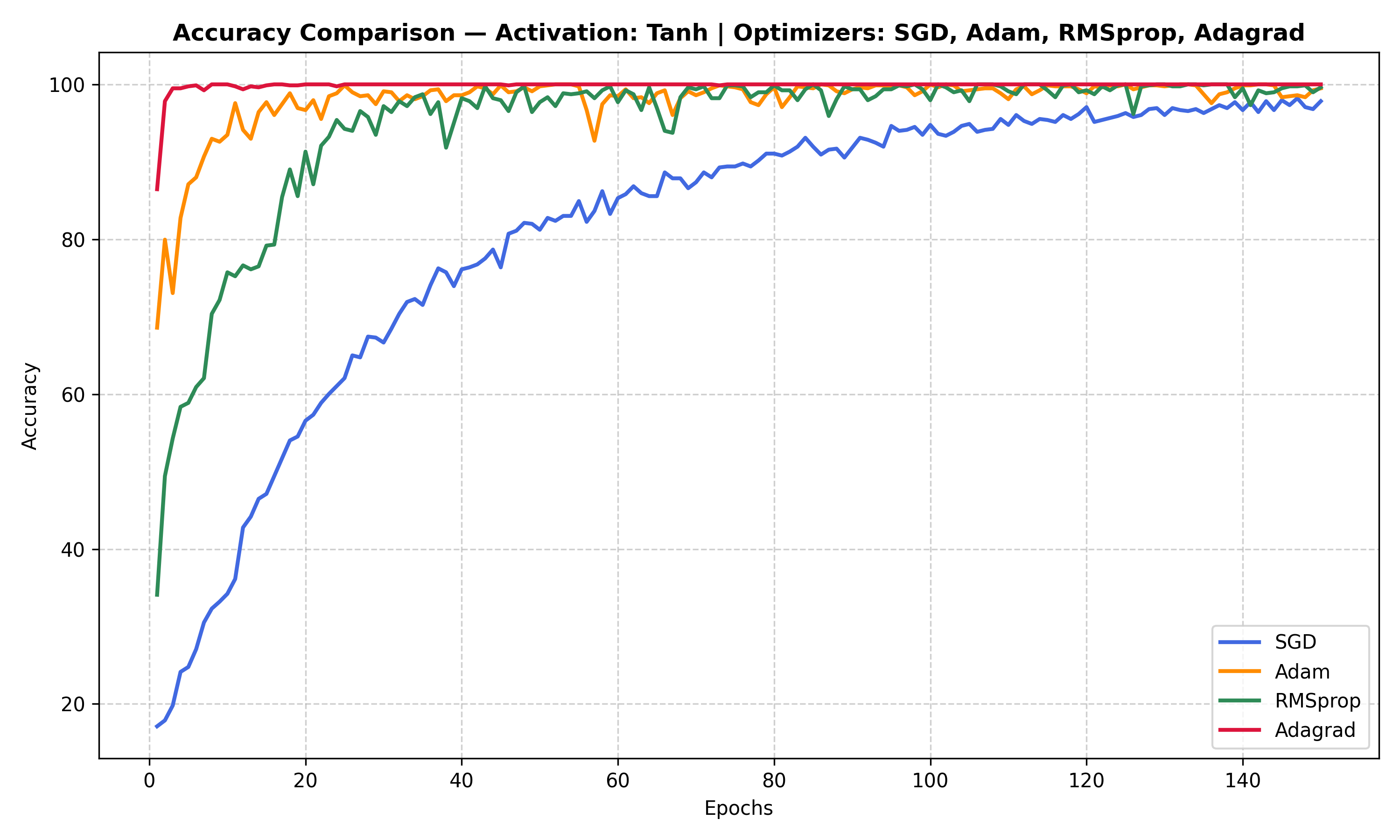}
\textbf{Figure 11(c).} Accuracy on UCF101 using Tanh
\end{minipage}
&
\begin{minipage}{\linewidth}
\centering
\includegraphics[width=\linewidth]{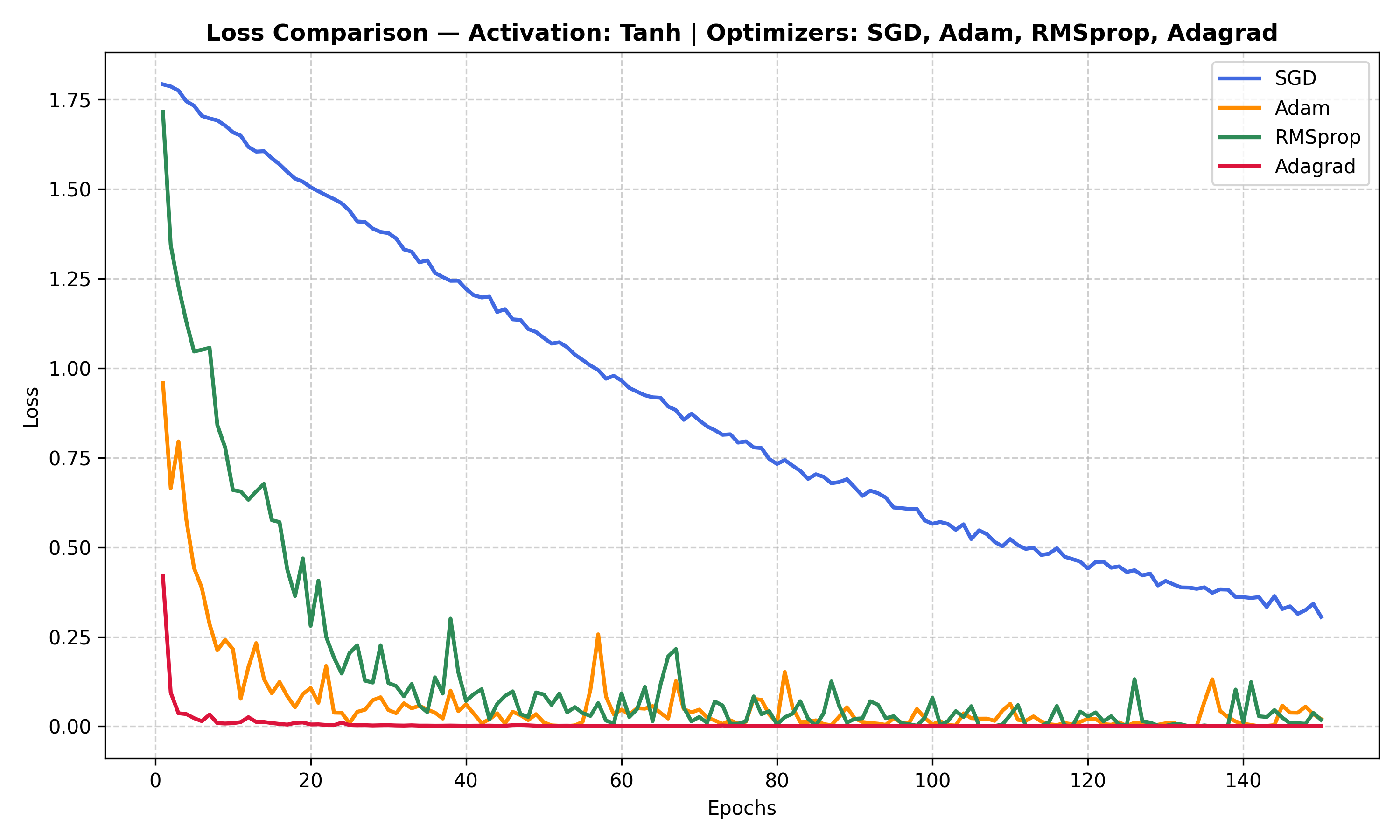}
\textbf{Figure 11(d).} Loss on UCF101 using Tanh 
\end{minipage}
\\ \hline

\end{longtable}
\end{center}

\twocolumn

% ============================================================
%                END OF TABLE FIGURE SECTION
% ============================================================

In this study, the performance of the ConvLSTM and BiLSTM models was evaluated across different AFs and MOs on both the UCF101 and HMDB51 datasets. Across all experiments, ConvLSTM demonstrated more stable and consistently stronger performance, especially when combined with RMSProp. With ReLU, Sigmoid, and even Tanh, RMSProp enabled ConvLSTM to reach near-perfect accuracy almost 100 percent on both datasets. Adagrad also produced excellent results, with all activation functions achieving close to 99 \% accuracy for UCF101 and HMDB51. Adam, although slightly more variable, still performed strongly, particularly with ReLU and Sigmoid, producing accuracy values above 95–98 percent. In contrast, SGD consistently lagged, reaching almost 60 \% accuracy depending on the activation function, making it the weakest optimizer for ConvLSTM in this experiment.

For the BiLSTM model, the trend was similar but with slightly greater variation between datasets. On UCF101, BiLSTM achieved its highest performance with Adagrad, especially when paired with ReLU, Sigmoid, or Tanh frequently reaching almost 99 percent accuracy. RMSProp and Adam also provided reliable results on UCF101, maintaining accuracy levels above 98\%. On the more challenging HMDB51 dataset, BiLSTM showed noticeable performance drops. As observed in ConvLSTM, SGD produced the lowest accuracy across nearly all activation functions for BiLSTM on both datasets.
	
ConvLSTM is better suited for HMDB51, where the videos often contain clutter, irregular motion, and noise. Its simpler structure allows it to generalize well under such conditions. On the other hand, BiLSTM performs exceptionally on the cleaner UCF101 dataset, where its deeper temporal modeling benefits from more consistent motion patterns. Because both models rely heavily on spatial features and less on long-range temporal cues, they achieve higher accuracy on UCF101, while HMDB51 being more temporally complex remains more challenging.The complete accuracy results for both models are provided in Table \ref{tab:convLSTM} and Table \ref{tab:BiLSTM}. 

\begin{table}[htbp]
\centering
% \tiny
\begin{minipage}{0.47\textwidth}
\centering
\caption{ConvLSTM Model Performance}
\label{tab:convLSTM}
\begin{tabular}{|c|c|c|c|c|}
\hline
\textbf{SN} & \textbf{AF} & \textbf{MO} & \textbf{HMDB51} & \textbf{UCF101} \\ \hline
1 & ReLU & SGD & 85.11\% & 98.99\% \\ \hline
2 & ReLU & Adam & 99.14\% & 99.15\% \\ \hline
3 & ReLU & RMSProp & 99.24\% & 99.87\% \\ \hline
4 & ReLU & Adagrad & 78.74\% & 99.11\% \\ \hline
5 & Sigmoid & SGD & 84.89\% & 72.27\% \\ \hline
6 & Sigmoid & Adam & 99.31\% & 99.68\% \\ \hline
7 & Sigmoid & RMSProp & 99.57\% & 99.64\% \\ \hline
8 & Sigmoid & Adagrad & 85.71\% & 98.07\% \\ \hline
9 & Tanh & SGD & 70.34\% & 99.03\% \\ \hline
10 & Tanh & Adam & 99.36\% & 99.09\% \\ \hline
11 & Tanh & RMSProp & 99.14\% & 99.85\% \\ \hline
12 & Tanh & Adagrad & 85.21\% & 98.78\% \\ \hline
\end{tabular}
\end{minipage}
\hfill
\end{table}

Table \ref{tab:convLSTM} presents the accuracy of the ConvLSTM model on both the UCF101 and HMDB51 datasets using different activation function and optimizer combinations. The results show that ConvLSTM delivers highly stable and consistent performance across all experimented settings. RMSProp and Adagrad emerge as the most effective optimizers, achieving near-perfect accuracy with ReLU, Sigmoid, and Tanh on both datasets. SGD consistently provides the lowest accuracy, indicating its limited effectiveness for this model. Overall, ConvLSTM performs reliably on both datasets, with particularly strong generalization on HMDB51.

\begin{table}[htbp]
\centering
% \tiny
\begin{minipage}{0.47\textwidth}
\centering
\caption{BiLSTM Model Performance}
\label{tab:BiLSTM}
\begin{tabular}{|c|c|c|c|c|}
\hline
\textbf{SN} & \textbf{AF} & \textbf{MO} & \textbf{HMDB51} & \textbf{UCF101} \\ \hline
1 & ReLU & SGD & 54.98\% & 98.09\% \\ \hline
2 & ReLU & Adam & 57.37\% & 99.15\% \\ \hline
3 & ReLU & RMSProp & 56.97\% & 99.35\% \\ \hline
4 & ReLU & Adagrad & 60.96\% & 99.77\% \\ \hline
5 & Sigmoid & SGD & 55.78\% & 99.36\% \\ \hline
6 & Sigmoid & Adam & 61.35\% & 99.31\% \\ \hline
7 & Sigmoid & RMSProp & 58.57\% & 98.87\% \\ \hline
8 & Sigmoid & Adagrad & 62.95\% & 99.87\% \\ \hline
9 & Tanh & SGD & 57.37\% & 96.82\% \\ \hline
10 & Tanh & Adam & 61.75\% & 99.03\% \\ \hline
11 & Tanh & RMSProp & 61.75\% & 99.77\% \\ \hline
12 & Tanh & Adagrad & 62.95\% & 99.31\% \\ \hline
\end{tabular}
\end{minipage}
\end{table}

Table \ref{tab:BiLSTM} summarizes the BiLSTM model’s accuracy across different activation functions and optimizers for the UCF101 and HMDB51 datasets. BiLSTM achieves excellent performance on UCF101, especially when paired with Adagrad or RMSProp, frequently reaching close to 99 \% accuracy. Its accuracy decreases on HMDB51 due to the dataset’s complex and noisy temporal patterns. Similar to ConvLSTM, SGD remains the weakest optimizer across all activation functions. These findings highlight that BiLSTM is better suited for cleaner datasets like UCF101, while HMDB51 benefits more from simpler architectures.

\section{Comparative Performance of ConvLSTM and BiLSTM Models}

This section provides a detailed analysis of the performance of two DL models evaluated on a six-class subset of both datasets. The ConvLSTM model achieved up to 98 \% accuracy with specific configurations, substantially outperforming BiLSTM, particularly on HMDB51. The following subsections discuss architectural, training, and hyperparameter factors responsible for these performance differences. The analysis explores the architectural, training, and hyperparameter factors contributing to ConvLSTM’s superior performance and com-pares these results with prior studies on datasets, explaining why the proposed approach achieved higher accuracy.

\subsection{Comparative Analysis of Model Performance and Superior Accuracy}
ConvLSTM outperformed BiLSTM due to its ability to preserve spatial structure while simultaneously modeling temporal evolution, which is critical for motion-rich datasets like HMDB51. ConvLSTM integrates convolutional operations to maintain spatial features of video frames while using recurrent gates to capture temporal dynamics, allowing the model to learn motion transitions and visual cues effectively, which is essential for HAR tasks \cite{Ref46}. Furthermore, configurations using ReLU reduced gradient saturation, enabling deeper temporal modeling and faster, more stable convergence during training. The compatibility of ConvLSTM with adaptive optimizers such as Adam and RMSprop further enhanced its learning stability by dynamically adjusting learning rates, helping the model capture fine-grained motion differences across both datasets \cite{Ref47}. ConvLSTM demonstrated strong performance in recognizing complex motion patterns, including subtle movements and multi-stage temporal changes, thanks to its dual ability to process sequences while retaining spatial details. This combination of spatial-temporal learning makes ConvLSTM particularly suited for HAR tasks, as it effectively understands both what is happening and how it changes over time, a crucial aspect of real-life human actions.

\subsection{Analysis of BiLSTM Model Performance}
The BiLSTM model, trained from scratch with the same AF and MO combinations, achieved 60.00\% accuracy on the HMDB51 dataset but reached 98.00\% on UCF101. This lower performance on HMDB51 can be attributed to several factors. BiLSTM excels at modeling temporal dependencies but lacks a dedicated spatial feature extractor, making it less effective for tasks involving video frames, where spatial patterns are crucial for accurate recognition \cite{Ref46}. Additionally, BiLSTM was trained without pre-trained weights, which required it to learn both spatial and temporal features from scratch, putting it at a disadvantage, particularly when dealing with the limited size of the HMDB51 subset \cite{Ref47}. The use of Sigmoid and Tanh activation functions also contributed to vanishing gradient issues, which slowed learning and decreased accuracy, although the use of ReLU slightly alleviated this problem, it was not enough without pre-trained spatial features.
Furthermore, the small dataset size in HMDB51 limited BiLSTM's ability to learn complex temporal patterns, resulting in poor generalization. While BiLSTM processes sequences in both forward and backward directions, which helps capture the full context of actions like sitting down or standing up, it struggled with shorter or visually similar actions in HMDB51, which exhibited weaker temporal flow. Despite BiLSTM's effectiveness on UCF101, where actions are long and clearly defined, its performance in datasets with complex motion patterns remains limited. Additionally, limited research has examined the combined impact of AF and MO choices on recurrent video models, especially in medically motivated six-class subsets of datasets like HMDB51 and UCF101. This gap in the literature underscores the need for a comprehensive evaluation of how such combinations influence model performance, particularly in optimizing HAR systems for real-world applications. We performed sensitivity checks to understand major contributors to performance. Removing temporal down-sampling reduced ConvLSTM accuracy by 3–5 \% due to longer sequences introducing noise. Omitting frame normalization increased training instability and validation loss, suggesting normalization is essential for convergence. Training/validation curves show no severe overfitting for ConvLSTM; BiLSTM shows modest overfitting on HMDB51, likely due to smaller intra-class temporal consistency.

\section{Conclusion \& Future Works}
This research presents a comprehensive evaluation of the impact of AF and MO on the performance of deep learning models in HAR. The study demonstrates that the ConvLSTM architecture, which integrates convolutional operations with LSTM-style gates, outperforms the BiLSTM model, particularly in capturing spatio-temporal features. ConvLSTM, when paired with optimizers like Adam or RMSprop and ReLU, achieves up to 99.00\% accuracy across datasets, underscoring its strong capability in complex HAR tasks, such as healthcare monitoring and surveillance. In contrast, BiLSTM exhibited reduced flexibility, achieving high accuracy only on specific datasets (UCF101), highlighting its limitations in handling noisy or challenging conditions like those in HMDB51.
The findings emphasize the importance of carefully selecting AF and MO combinations for optimizing model performance. ConvLSTM with ReLU and either Adam or RMSprop stands out as the most reliable configuration for robust, high-accuracy HAR systems. This work provides empirical insights into model configuration, showing that the right combination of activation function and optimizer can significantly improve convergence speed, classification accuracy, and real-world applicability in dynamic environments. BiLSTM, while effective in some cases, struggles with certain activity classes, especially those with complex motion dynamics, indicating that a hybrid spatio-temporal approach like ConvLSTM is more effective.

The study’s findings open several promising directions for future research. Adaptive AF and MO strategies can further improve model training efficiency and robustness, particularly for small or noisy datasets commonly encountered in medical or rehabilitation contexts. Future work could explore dynamic gating mechanisms or hybrid models that enhance the temporal and spatial feature extraction of ConvLSTM and BiLSTM. Incorporating attention mechanisms or lightweight transformer models could improve long-range dependency modeling without increasing computational cost, supporting real-time applications in surveillance and smart environments. Evaluating these models on larger, more diverse, and clinically relevant datasets, as well as in resource-constrained conditions, is crucial for developing scalable HAR systems. These advancements will enhance the real-world deployability of HAR models, particularly in healthcare, elderly care, and ambient assisted living contexts. The study is limited to public video datasets and a restricted set of activity classes, and future validation on clinical data and real-time deployment scenarios is essential to confirm the robustness and scalability of treal-worldnded configurations. Analyzing learning curves revealed that Adam and RMSprop converge faster due to adaptive learning-rate updates, while SGD exhibited slow and unstable convergence. ReLU allowed for more stable gradient flow, especially in deeDeveloping dynamic or hybrid gating mechanisms in ConvLSTM and BiLSTM may enhance temporal and spatial feature extraction, allowing models to detect subtle activity variations more accurately, which is beneficial for fall detection, elderly care monitoring, and physical therapy assessment.lications, particularly in healthcare and surveillance.

\section*{Acknowledgement}
Not applicable.

\section*{Funding Statement}
Not applicable.

\section*{Author Contributions}
Md. Ekramul Hamid led the research and supervised the project. Subrata Kumer Paul, Dewan Nafiul Islam Noor, and Abu Saleh Musa Miah designed the methodology; Subrata Kumer Paul also collected and processed data. Software was implemented by Rakhi Rani Paul, Subrata Kumer Paul, and Md. Ekramul Hamid. The draft was written by Subrata Kumer Paul, Abu Saleh Musa Miah, and Rakhi Rani Paul, with all authors contributing to revisions. Abu Saleh Musa Miah handled administration and funding, while Dewan Nafiul Islam Noor oversaw validation and review. All authors reviewed the results and approved the final version of the manuscript.

\section*{Availability of Data and Materials}
Not applicable.

\section*{Ethics Approval}
Not applicable.

\section*{Conflicts of Interest}
The authors declare no conflicts of interest regarding the present study.

%%%%%%%%%%%%%%%%%%%%%%%%%%%%%%%%%%%%%%%%%%%%%%%%%%%%%%%%%%%%%%%%%%%%%%%%%%%%%%%%%%%%%%%%
%% ATENCION AUTORES: ESTA SECCION ES OBLIGATORIA
%%%%%%%%%%%%%%%%%%%%%%%%%%%%%%%%%%%%%%%%%%%%%%%%%%%%%%%%%%%%%%%%%%%%%%%%%%%%%%%%%%%%%%%%



%% References


\bibliographystyle{elsarticle-num} 
\bibliography{bibtext}

\end{document}